\pdfoutput=1
\PassOptionsToPackage{table}{xcolor}
\documentclass[11pt]{article}

\usepackage[preprint]{acl}

\usepackage{times}
\usepackage{latexsym}
\usepackage{soul} 
\definecolor{TeacherColor}{HTML}{B3CDE3}
\definecolor{StudentColor}{HTML}{CCEBC5}

\newcommand{\highlightteacher}[1]{\sethlcolor{TeacherColor}\hl{#1}}
\newcommand{\highlightstudent}[1]{\sethlcolor{StudentColor}\hl{#1}}

\usepackage[T1]{fontenc}

\usepackage[utf8]{inputenc}

\usepackage{microtype}

\usepackage{inconsolata}

\usepackage{graphicx}
\usepackage{subcaption}
\usepackage{amsmath}
\usepackage{amssymb}
\usepackage{pifont}
\usepackage{booktabs} 
\usepackage[linesnumbered,ruled,vlined]{algorithm2e}
\usepackage{multirow}
\usepackage{float}

\newcommand{\myparagraph}[1]{\noindent\textbf{#1}}
\newcommand{\cmark}{\ding{51}}%
\newcommand{\xmark}{\ding{55}}%

\title{Lillama: Large Language Models Compression via Low-Rank Feature Distillation}

\begin{NoHyper}

\author{Yaya Sy \\ \texttt{yaya.sy@loria.fr} \And Christophe Cerisara \\ \texttt{christophe.cerisara@loria.fr}  \And Irina Illina \\ \texttt{irina.illina@loria.fr} \AND \normalfont{LORIA, CNRS, Nancy, France}
 }
\end{NoHyper}

\begin{document}
\maketitle

\begin{abstract}
Current LLM structured pruning methods typically involve two steps: (1) compression with calibration data and (2) costly continued pretraining on billions of tokens to recover lost performance. This second step is necessary as the first significantly impacts model accuracy. Prior research suggests pretrained Transformer weights aren’t inherently low-rank, unlike their activations, which may explain this drop. Based on this observation, we propose \textbf{Lillama}, a compression method that locally distills activations with low-rank weights. Using SVD for initialization and a joint loss combining teacher and student activations, we accelerate convergence and reduce memory use with local gradient updates. \textbf{Lillama} compresses Mixtral-8x7B within minutes on a single A100 GPU, removing 10 billion parameters while retaining over 95\% of its original performance. Phi-2 3B can be compressed by 40\% with just 13 million calibration tokens, resulting in a small model that competes with recent models of similar size. The method generalizes well to non-transformer architectures, compressing Mamba-3B by 20\% while maintaining 99\% performance\footnote{Code available at \url{https://github.com/yaya-sy/lillama}}.
\end{abstract}

\begin{figure}[ht]
        \centering
        \includegraphics[width=\linewidth]{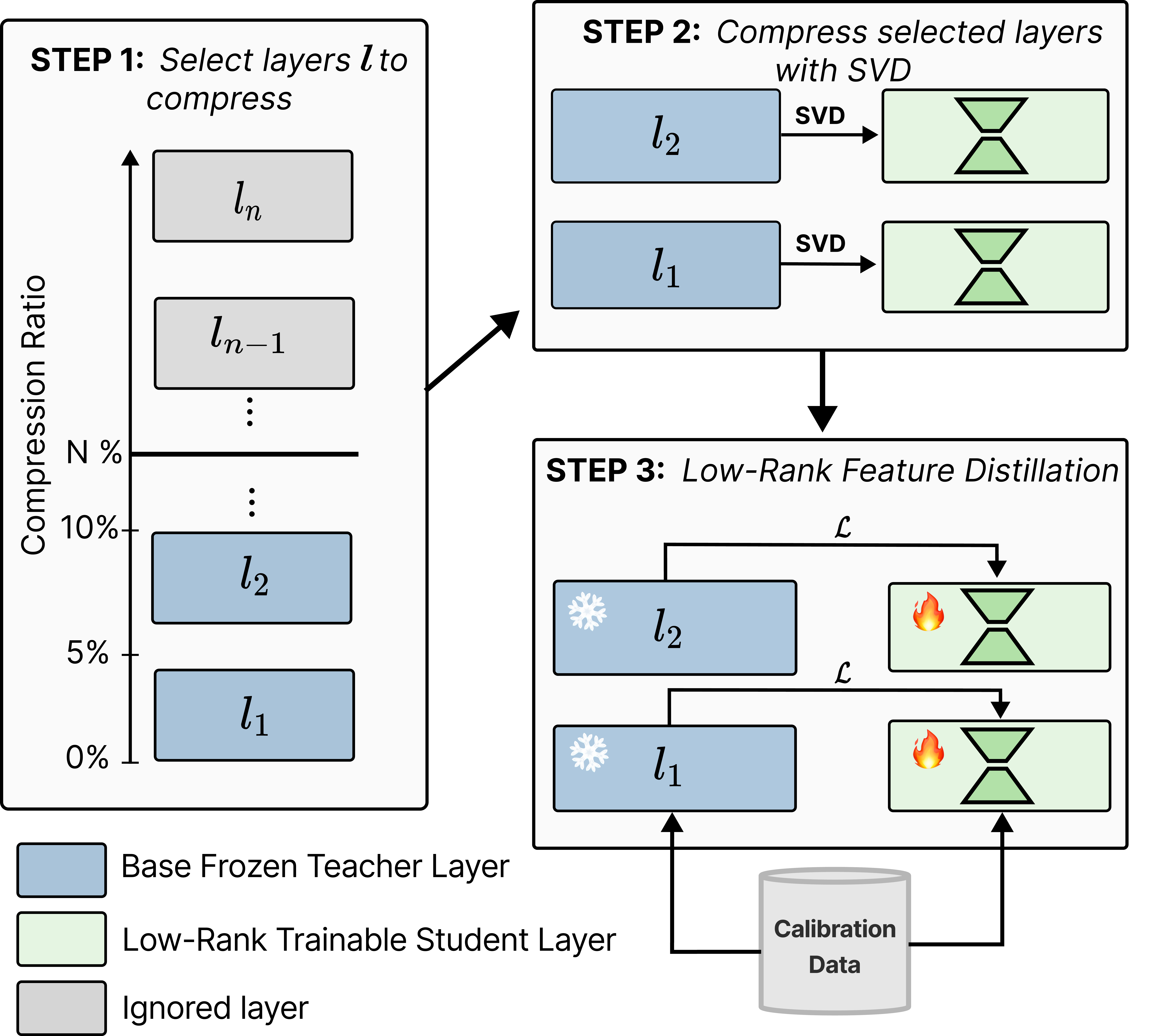}
        \caption{\textbf{Lillama approach}: \textbf{STEP 1} selects layers to compress for a target compression ratio (e.g., N\%) using various strategies (see Section \ref{sec:algo}). \textbf{STEP 2} compresses and initializes the chosen parameters via SVD. \textbf{STEP 3} distills the low-rank weights with a small calibration dataset.}
        \label{fig:overview}
\end{figure}

\section{Introduction}
\begin{table*}[h!]
\resizebox{\textwidth}{!}{%
\setlength{\tabcolsep}{3pt}
\begin{tabular}{c|c|c|c|c|c}
\toprule
\textbf{Method} & \textbf{No Custom Kernel} & \textbf{No Ampere-only} & \textbf{Memory Gain} & \textbf{Is One-Shot} & \textbf{Calib. Data} \\  \hline
SparseGPT \citep{frantar2023sparsegptmassivelanguagemodels} & \xmark & \xmark & \xmark & \cmark & - \\
Wanda \citep{wanda} & \xmark & \xmark & \xmark & \cmark & - \\
Teal \citep{TEAL} & \xmark & \xmark & \xmark & \cmark & - \\
ShearedLLaMA \citep{xia2024sheared} & \cmark & \cmark & \cmark & \xmark & 52B\\ 
Minitron \citep{minitron} & \cmark & \cmark & \cmark & \xmark & 94B \\
\rowcolor{gray!15} \textbf{Lillama} & \cmark & \cmark & \cmark & \cmark & \textbf{0.013B} \\
\bottomrule
\end{tabular}}
\caption{Positioning of our approach within the compression literature. \textbf{No Custom Kernel} indicates that a custom GPU kernel is not required to observe speedup and \textbf{No Ampere-only} is whether the method requires Ampere GPU to achieve speedup. \textbf{Calib. Data} is the total number of tokens in billions used for compression.}
\label{tab:litt}
\end{table*}
Compressing Large Language Models is a crucial challenge that can be achieved with several complementary approaches: quantization, pruning, compression, and distillation. We propose a new one-shot compression method that locally distills low-rank weights and satisfies the three objectives:
\begin{enumerate}
    \item The compression algorithm must be compute-efficient, e.g., it should require minimal computational resources and run in a reasonable amount of time. For instance, the iterative magnitude pruning algorithm as done in \citet{frankle2019lottery}, is too costly to be applied to large-scale LLMs.
    \item It must achieve fast convergence using as little training data as possible; this objective is often an issue with many model pruning methods that often require significant retraining after pruning to recover the lost performance.
    \item The method should not cause a significant drop in performance; ideally, all the capabilities of the original LLM should remain intact.
\end{enumerate}
To address the first constraint, we compress the model with a local distillation objective, as opposed to a global objective that is more costly. We further reduce the cost by compressing and distilling only a subset of all layers.
About the second constraint, we initialize the compressed layers with Singular Value Decomposition (SVD), which enables 
reducing the required number of gradient steps as well as the calibration dataset size. We further improve convergence by combining Teacher and Student activations with a joint distillation loss. 
We experimentally show the robustness of our approach by compressing several state-of-the-art small and large Transformers, Mixture-of-Experts, and Mamba LLMs.
Figure~\ref{fig:overview} gives an intuitive overview of the proposed method, which is detailed in Section~\ref{sec:method} and validated in Section~\ref{sec:xp}.

\section{Related Works}
We discuss next previous compression methods, highlight their limitations, and explain how our approach addresses them. Table~\ref{tab:litt} positions our approach to recent compression methods.

\noindent{\bf{Pruning}} methods remove unimportant weights in the pre-trained model~\citep{NIPS1989_6c9882bb,han2015learning}. Structured Pruning removes entire groups of parameters, which results in a smaller and faster model~\citep{xia2024sheared,ma2023llmpruner}. \citet{ma2023llmpruner} propose a new gradient-based criterion to eliminate substructures in LLMs, while \citet{xia2024sheared} use a joint loss combining a pruning mask loss with the language modeling loss. However, optimizing these criteria can be computationally intensive. For example, the pruning step of Sheared-LLaMA~\citep{xia_sheared_2023} is 5x expensive compared to standard LM training, according to the authors. In contrast, thanks to the local gradient updates, our approach is computationally efficient, allowing us to compress a 47B model within minutes on a single A100 GPU. Regarding unstructured pruning, these methods do not provide any gains in terms of memory or speedup, at least with current algorithmic implementations. Semi-structured pruning (e.g., 2:4 and 4:8)~\citep{wanda, frantar2023sparsegptmassivelanguagemodels, TEAL} does not lead to memory gain but can speed up processing on kernels optimized for such matrix structures. On the other hand, our method, which directly shrinks matrices, saves memory across all hardware and leads to speed up, as fewer computations are performed.

\begin{figure}[t]
    \centering
        \begin{subfigure}[b]{0.15\textwidth}
            \centering
            \includegraphics[width=\textwidth]{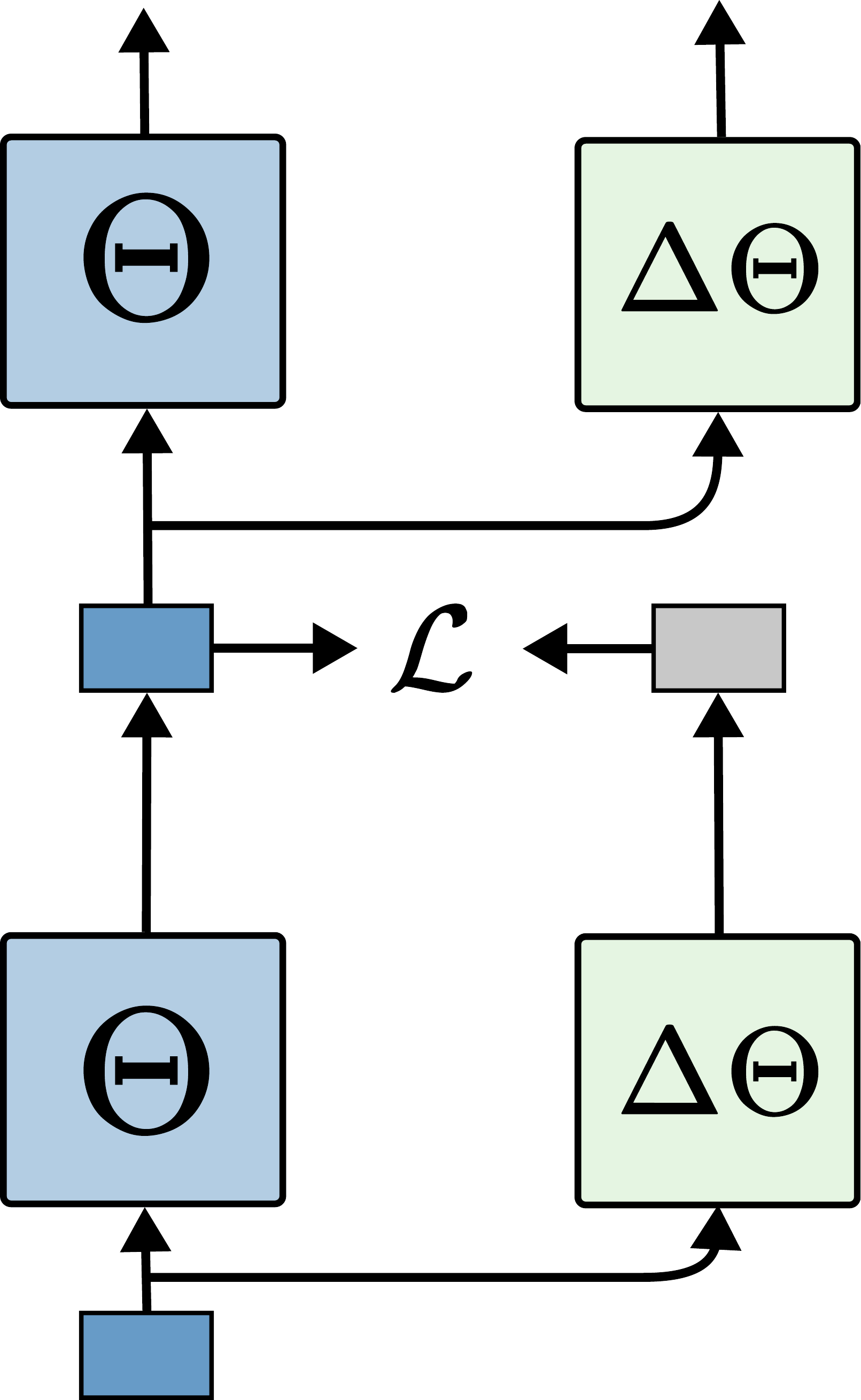}
            \caption{\scriptsize Teacher}
            \label{fig:teacher}
        \end{subfigure}
        \hfill
        \begin{subfigure}[b]{0.15\textwidth}
            \centering
            \includegraphics[width=\textwidth]{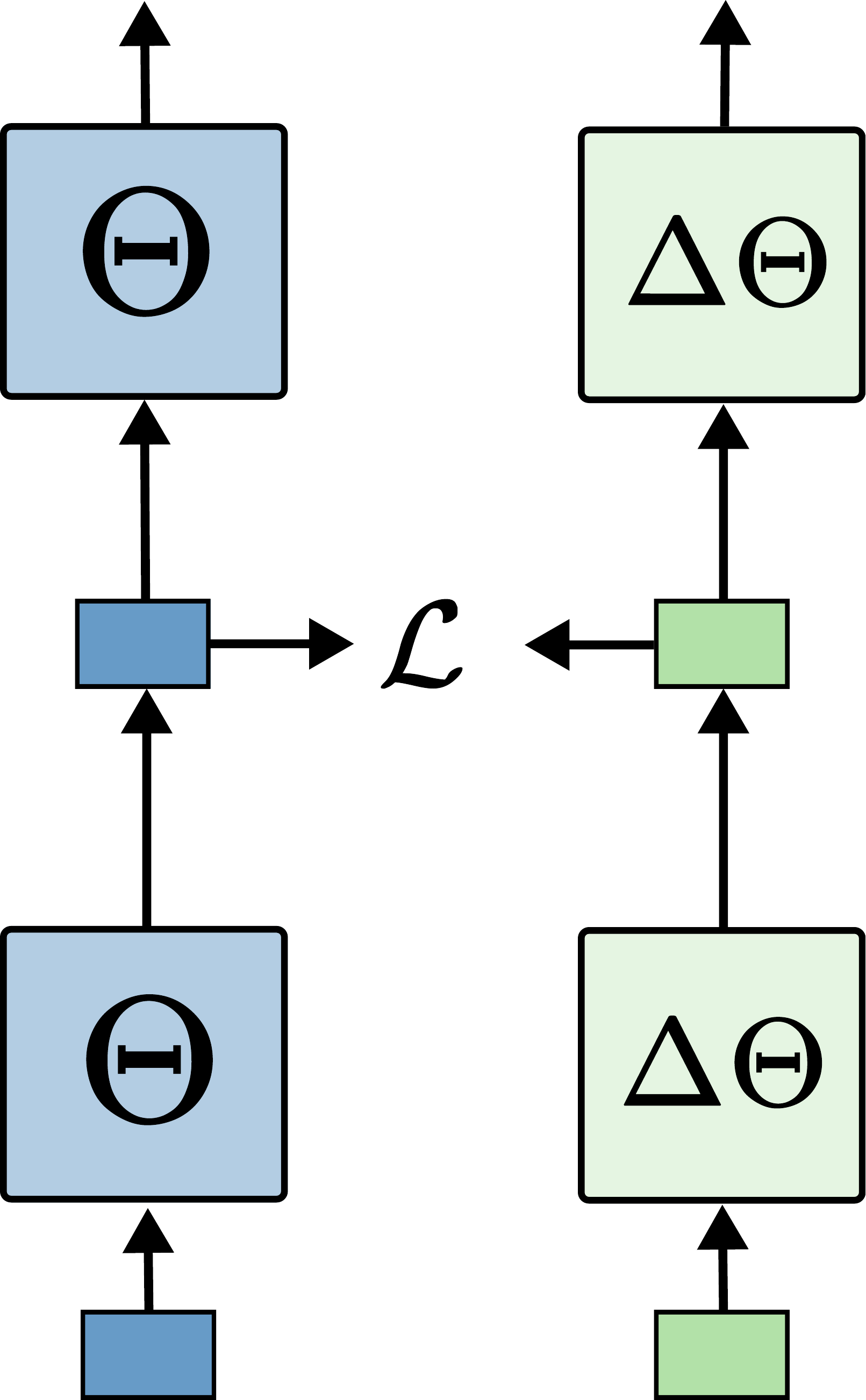}
            \caption{\scriptsize Student}
            \label{fig:student}
        \end{subfigure}
        \hfill
        \begin{subfigure}[b]{0.15\textwidth}
            \centering
            \includegraphics[width=\textwidth]{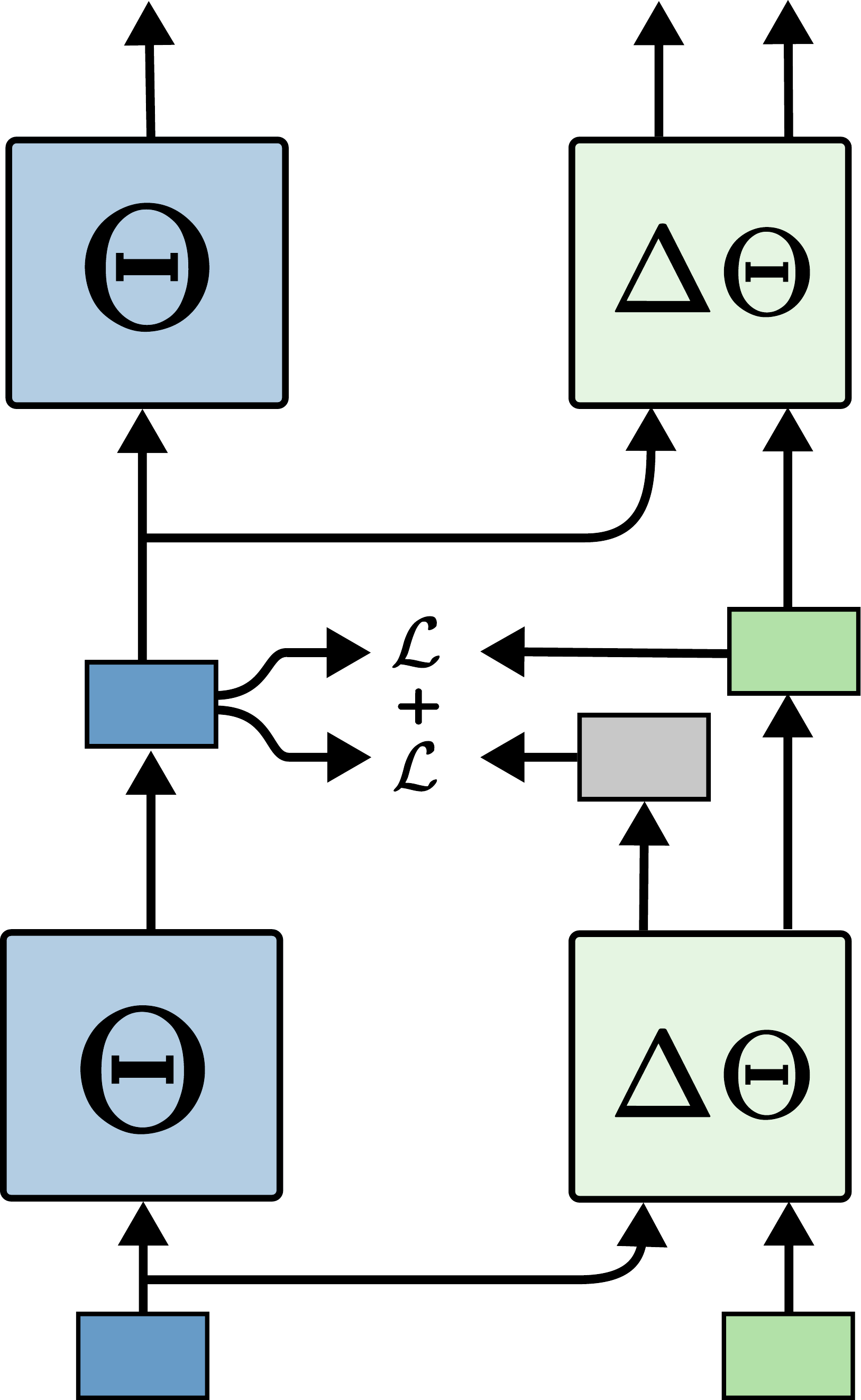}
            \caption{\scriptsize Teacher+Student}
            \label{fig:teacher_student}
        \end{subfigure}
        \caption{\textbf{Accelerating convergence by learning from \highlightteacher{Teacher} and \highlightstudent{Student} activations through a joint loss}. We propose to study the effect of three distillation strategies: \textbf{(a) Teacher}: the input to the compressed student layer comes from the output of the previous teacher layer; \textbf{(b) Student}: the input to the compressed student layer comes from the output of the previous student layer; \textbf{(c) Teacher+Student}: the compressed student layer receives both the output of the previous teacher layer and the output of the previous student layer, and the loss is the sum of two losses.}
    \label{fig:distillaton}
\end{figure}

\noindent{\bf Low-Rank Decomposition} compresses a model by approximating its pre-trained matrices with lower-dimension ones. Based on the observation by \citet{li2018measuring} that Neural Networks have lower intrinsic dimensions, \citet{aghajanyan2020intrinsic} show that Transformer language models also require lower intrinsic dimensions depending on the language task. However, \citet{yu_compressing_2023, chen_drone} show that the weights in some pretrained Transformer models are full-rank compared to their activations and propose an activation-aware compression method. Our work is related to these approaches, as we also decompose the matrices of the models. However, we apply a lightweight layer-wise feature distillation objective to better recover from compression and propose different distillation strategies to accelerate convergence and improve performance.

\noindent{\bf Distillation.} In Knowledge Distillation (KD), a small student model learns to reproduce the outputs of a larger pre-trained model. Standard KD is data-intensive when the student model's parameters are randomly initialized. \citet{gemmateam2024gemma2improvingopen} recently distilled Gemma-2 9B from a larger model using 8 trillion tokens, which is as costly as a standard pretraining stage. Another approach is layer-wise distillation, which can be more suitable in some cases, as intermediate layers may contain certain information, such as speaker identity or phonemes in pretrained speech models~\citep{baevski2020wav2vec20frameworkselfsupervised, pasad2022layerwiseanalysisselfsupervisedspeech, chang2022distilhubertspeechrepresentationlearning, chang2021explorationselfsupervisedpretrainedrepresentations}. For example, TinyBERT~\citep{jiao_tinybert_2020} initializes the student model by removing layers from the teacher and then distills the student on task-specific knowledge, combining layer-wise and global distillation losses, which took 3.5 days to train. This is costly for LLMs. Instead of removing layers, we leverage the observation that layer activations are low-rank and approximate them with fewer parameters. Additionally, we compress the models to remain generalist rather than task-specific.

\section{Background: Low-Rank Approximation of Transformer Models}

\subsection{Low-Rank Approximation from Weights.} 
Given $W \in \mathbb{R}^{d_{1} \times d_{2}}$, a pretrained weight matrix, and an example $x \in \mathbb{R}^{d_{2}}$, the activations $y \in \mathbb{R}^{d_{1}}$ are computed as $y = Wx$. It is possible to reduce the computations in this operation by using a low-rank approximation of $W$: $y \simeq \Delta Wx = ABx$ with $\Delta W = AB$ the low-rank decomposition of $W$, composed of $A \in \mathbb{R}^{d_{1} \times r}$ and $B \in \mathbb{R}^{r \times d_{2}}$. When the rank $r$ is small enough, the number of parameters $r(d_1+d_2)$ in the low-rank equation is smaller than $d_1d_2$ in the full-rank equation. Estimating the low-rank matrix $\Delta W = AB$ can be formulated as a minimization problem to find the low-rank $\Delta W$ that best approximates $W$:

\begingroup
\setlength{\abovedisplayskip}{0pt}
\setlength{\belowdisplayskip}{3pt}
{\small{
\begin{align}
    \label{eq:lr_weight}
    \widehat{\Delta W} &= \underset{{\Delta W}}{\mathrm{argmin}} \;\; \|W - {\Delta W}\|_{F}
\end{align}
}}
\endgroup
It is well-known that this minimization problem can be approached using SVD, a low-rank approximation method that offers the optimal $r-$rank approximation of a given matrix with regard to the Frobenius norm $\|\cdot\|_{F}$.
SVD approximates a matrix $W$ into three matrices: $W = USV^{T}$, where $S\in \mathbb{R}^{d_{1} \times d_{2}}$ is a diagonal matrix containing the singular values of $W$ sorted in descending order, $U \in \mathbb{R}^{d_{1} \times d_{1}}$ and $V \in \mathbb{R}^{d_{2} \times d_{2}}$ are orthonormal matrices. The optimal low-rank approximation of $W \in \mathbb{R}^{d_{1} \times d_{2}}$ can be obtained by keeping the first $r$ singular values, with $r < min(d_{1}, d_{2})$:

\begingroup
\setlength{\abovedisplayskip}{0pt}
\setlength{\belowdisplayskip}{3pt}
{\small{
\begin{align}
    \label{eq:svd}
    \widehat{\Delta W} &= (U_{:, :r}S_{:r, :r})V_{:r,:} = AB
\end{align}
}}
\endgroup
with $A=U_{:,:r}S_{:r,:r}$ and $B=V_{:r,:}$ and the notation $_{:a, :b}$ refers to the slicing operation. 

\subsection{Low-Rank Approximation from Feature}
\label{ssec:pca}

Previous studies~\citep{chen_drone, yu_compressing_2023} have shown that the activations (i.e., \textit{features}) of pretrained transformers, are more low-rank than the weights. Recently, \citet{TEAL} show that transformer activations can be sparsified up to 60\% without too much drop in performance. In Appendix \ref{ssec:srank}, we also show that the activations of the transformer layers are more low-rank than the weights. All these results suggest transformer activations can be approximated with fewer parameters. Now, let's examine how previous works \citep{kaushal_lord_2023, chen_drone, yu_compressing_2023} have approached this.

Let $\mathcal{D} = \{x_i \in \mathbb{R}^{d_{2}}\}_{1\leq i\leq N}$ be a calibration dataset,
and $W \in \mathbb{R}^{d_{1} \times d_{2}}$ the parameters of a linear layer.
Finding the low-rank matrix $\Delta W$ that best reproduces the activations $\{y=Wx\}~~~ \forall x\in \mathcal{D}$ involves solving:

\begingroup
\setlength{\abovedisplayskip}{0pt}
\setlength{\belowdisplayskip}{3pt}
{\small{
\begin{align}
    \label{eq:activations_lr}
    \widehat{\Delta W} &= \underset{{\Delta W}}{\mathrm{argmin}} \;\; \frac{1}{N}\sum\limits_{x \in \mathcal{D}}\|Wx - {\Delta Wx}\|_{F}
\end{align}
}}
\endgroup
An analytic solution to this minimization problem is the eigendecomposition of the covariance matrix of the activations. We first collect all activations $\mathcal{Y}=\{y=Wx\}_{\forall x\in \mathcal{D}}$, then
the covariance matrix of the activations can be estimated as {\small{$\Sigma = \underset{y \in \mathcal{Y}}{\mathbb{E}}\left[yy^T\right] - \mathbb{E}[y]\mathbb{E}[y]^T$}}
with $\Sigma \in \mathbb{R}^{d_{1} \times d_{1}}$. Since $\Sigma$ is a diagonalizable matrix, we can apply its eigendecomposition: $\Sigma = USU^{T}$, where $U \in \mathbb{R}^{d_{1} \times d_1}$ contains the eigenvectors of $\Sigma$ and $S \in \mathbb{R}^{d_{1} \times d_{1}}$ is the diagonal matrix containing the eigenvalues sorted in decreasing order. As for Eq-\ref{eq:svd}, we can only keep the eigenvectors corresponding to the $r$ largest eigenvalues, which gives us $A = U_{:, :r}$ and $B =  U_{:, :r}^{T}W$.
Compared to the low-rank matrices in Eq-\ref{eq:svd}, $A$ and $B$ weights here learned to reproduce the activations of the base weight $W$, thanks to the minimization objective in Eq-\ref{eq:activations_lr}. In the next section, we highlight the limitations of this approach and introduce our proposed method.

\section{Proposed Approach}
\label{sec:method}
We identify and handle next three potential limitations of the approach presented in Section~\ref{ssec:pca}.

\noindent{\bf Non-linear Feature Approximation.}
The previous analytical solution for Eq-\ref{eq:activations_lr} is limited to activations from linear layers.
To generalize to non-linear modules, we first observe that Eq-\ref{eq:activations_lr} can be seen as a feature distillation objective that may be optimized numerically by gradient descent rather than analytically by eigendecomposition. Given the input batch {\small{$X \in \mathbb{R}^{d \times b}$}} of $b$ examples, we denote the output activations of the {\small{$i^{th}$}} Teacher module {\small{$\mathcal{T}^{(i)}$}} as 

\begingroup
\setlength{\abovedisplayskip}{0pt}
\setlength{\belowdisplayskip}{3pt}
{\small{
\begin{align}
    \label{eq:teacher_fw}
    Y^{(i)}= \mathcal{T}^{(i)}(X; \Theta^{(i)})
\end{align}
}}
\endgroup
where $ \Theta^{(i)}$ are the original pretrained matrices of the $i^{th}$ Teacher, and $Y^{(i)} \in \mathbb{R}^{d \times b}$ its output activations. Similarly, we note the output activations of the $i^{th}$ Student module $\mathcal{S}^{(i)}$ as :

\begingroup
\setlength{\abovedisplayskip}{0pt}
\setlength{\belowdisplayskip}{3pt}
{\small{
\begin{align}
    \label{eq:student_fw}
    \widehat{Y}^{(i)}= \mathcal{S}^{(i)}(X; \Delta \Theta^{(i)})
\end{align}
}}
\endgroup
where the Student module $\mathcal{S}^{(i)}$ is parametrized with the low-rank matrices $\Delta \Theta^{(i)}$ and $\widehat{Y}^{(i)} \in \mathbb{R}^{d \times b}$ are the output activations. We can express the distillation objective of the $i^{th}$ Student module as:

\begingroup
\setlength{\abovedisplayskip}{0pt}
\setlength{\belowdisplayskip}{3pt}
{\small{
\begin{align}
    \label{eq:distill}
    \widehat{\Delta  \Theta^{(i)}} &= \underset{{\Delta  \Theta^{(i)}}}{\mathrm{argmin}} \;\; \mathcal{L}^{(i)}(Y^{(i)}, \; \widehat{Y}^{(i)})
\end{align}
}}
\endgroup
where {\small{$\widehat{\Delta  \Theta^{(i)}}$}} are the estimated low-rank matrices of the $i^{th}$ Student module $\mathcal{S}^{(i)}$ and $\mathcal{L}^{(i)}(Y^{(i)}, \; \widehat{Y}^{(i)})$ is the loss measuring the distance between the activations of the Teacher and the Student.
We opted for the same loss as \citet{chang2022distilhubertspeechrepresentationlearning} because we observed that the $\ell_{1}$ loss yields instabilities:

\begingroup
\setlength{\abovedisplayskip}{0pt}
\setlength{\belowdisplayskip}{3pt}
{\footnotesize{
\begin{align}
    & \mathcal{L}^{(i)} = \mathcal{L}^{(i)}_{\ell_1} + \mathcal{L}^{(i)}_{\cos} \\
    & = \sum_{t=1}^{b} \left[ \frac{1}{D} \left\| Y^{(i)}_{t} - \widehat{Y}^{(i)}_{t} \right\|_1 
    - \log \sigma \left( \cos \left( Y^{(i)}_{t}, \widehat{Y}^{(i)}_{t} \right) \right) \right] \notag
\end{align}
}}
\endgroup
where $D$ is the hidden vectors dimension, $\sigma$ is the sigmoid activation and cos$(\cdot, \cdot)$ is the cosine similarity. We propose to approximate the low-rank weights of the Students through Gradient Descent. In the linear case, this should converge towards the eigendecomposition solution described in Section~\ref{ssec:pca}.
However, as we show later, we extend the distillation process to non-linear modules, which requires numerical optimization. We also show that initializing the Student's low-rank parameters with SVD, rather than randomly, improves convergence.

\myparagraph{Beyond Teacher-only Activations.} The second potential limitation of the approach described in Section \ref{ssec:pca} and equations \ref{eq:teacher_fw} and \ref{eq:student_fw} is that the Student and the Teacher modules take the same input $X$, which is the output of the precedent layer of the Teacher. This means that the Student is trained from the activations of the Teacher module only, which are not available at inference time. \citet{yu_compressing_2023} use this approach to approximate the Linear Layers of Transformer models, referring to it as \textit{Atomic Feature Mimicking}. Let $\widehat{Y}^{(i)}_{T} = \mathcal{S}^{(i)}(Y^{(i-1)}_{T}; \Delta \Theta^{(i)})$ be the output activations of the Student module when fed $Y^{(i-1)}_{T}$ the output activations of the Teacher $T$ at the previous layer $i-1$. The loss can then be written as:

\begingroup
\setlength{\abovedisplayskip}{0pt}
\setlength{\belowdisplayskip}{3pt}
{\small{
\begin{align}
    \label{eq:teacher_loss}
    \mathcal{L}^{(i)}_{T} &= \mathcal{L}^{(i)}(Y^{(i)}_{T}, \; \widehat{Y}^{(i)}_{T})
\end{align}
}}
\endgroup
where $Y^{T}_{i}$ are the gold activations of the current Teacher. This distillation procedure is illustrated in Figure~\hyperref[fig:teacher]{2a}. This approach can lead to a fast convergence as the Student modules learn from the gold and high-quality activations produced by the Teacher modules. However, it can cause a performance drop during inference as the Teacher activations will not be available. An alternative approach is to have the Student module take as input the activations of the previous Student modules rather than those of the Teacher modules:

\begingroup
\setlength{\abovedisplayskip}{0pt}
\setlength{\belowdisplayskip}{3pt}
{\small{
\begin{align}
    \label{eq:student_loss}
    \mathcal{L}^{(i)}_{S} &= \mathcal{L}^{(i)}(Y^{(i)}_{T}, \; \widehat{Y}^{(i)}_{S})
\end{align}
}}
\endgroup
where {\small{$\widehat{Y}^{(i)}_{S} =
\mathcal{S}^{(i)}(\widehat{Y}^{(i-1)}_{S}; \Delta \Theta^{(i)})$}} is the output
activations of the current Student module $S$ when taking as input the output
activations {\small{$\widehat{Y}^{(i-1)}_{S}$}} of the Student module of the
previous layer, and {\small{$Y^{(i)}_{T}$}} are the gold activations of the corresponding
Teacher module. This is the standard distillation and is illustrated in Figure~\hyperref[fig:student]{2b}.
However, although we found this loss performs
generally better than the previous one, it leads to a slow convergence due to
the errors in the activations of the Student modules. To address both issues,
we propose to combine the two losses:

\begingroup
\setlength{\abovedisplayskip}{0pt}
\setlength{\belowdisplayskip}{3pt}
{\small{
\begin{align}
    \label{eq:teacher+student}
    \mathcal{L}^{(i)}_{T+S} = \mathcal{L}^{(i)}_{T} + \mathcal{L}^{(i)}_{S}
\end{align}
}}
\endgroup
This distillation approach is illustrated in Figure~\hyperref[fig:teacher_student]{2c}. We don't introduce any hyperparameter for controlling the importance of one loss over another in the joint loss. We leave this for future work.

\myparagraph{Module-wise Distillation.} Finally, we observe that the functions
$\mathcal{S}$ can be any other module in the neural network than linear
layers: any sub-part of the model that outputs low-rank activations can be
compressed with our method. In our experiments, we distill at the Transformer
layer level and leave comparisons with other distillation levels for future
work. We also experiment with other models than pretrained
Transformer-based language models.

\section{Fast and Memory-Efficient Compression}
\label{sec:algo}
\begin{table*}[t]
    \resizebox{\textwidth}{!}{%
    \setlength{\tabcolsep}{3pt}
    \begin{tabular}{cccccccccccc}
        \toprule
        \textbf{Model} & \textbf{Reduction} & \textbf{TruQA} & \textbf{SIQA} & \textbf{LogiQA} & \textbf{WinoG} & \textbf{ARC-E} & \textbf{ARC-C} & \textbf{BoolQ} & \textbf{PIQA} & \textbf{OBQA} & \textbf{Average} \\
        \midrule 
        \multirow{2}*{\centering Phi-3 14B} 
        & 0\% & 57.63 & 57.06 & 37.94 & 76.01 & 81.36 & 61.60 & 88.56 & 81.34 & 50.40 & \textbf{65.77} \\
        & 20\% & 57.88 & 50.26 & 34.10 & 72.53 & 83.54 & 59.22 & 85.32 & 80.30 & 49.20 & \textbf{63.59} \\
        \midrule
        \multirow{2}*{\centering Mixtral-8x7B-v0.1 47B} 
        & 0\% & 48.58 & 49.54 & 33.18 & 76.64 & 83.50 & 60.07 & 85.23 & 83.41 & 47.00 & \textbf{63.02} \\
        & 20\% & 44.78 & 47.85 & 30.41 & 72.85 & 80.81 & 56.74 & 81.41 & 80.30 & 46.60 & \textbf{60.19} \\
        \midrule
        \multirow{2}*{\centering Phi-2 3B} 
        & 0\% & 44.40 & 55.42 & 30.57 & 76.16 & 78.16 & 54.18 & 83.21 & 79.11 & 51.20 & \textbf{61.38} \\
        & 20\% & 46.04 & 52.56 & 29.80 & 71.51 & 72.81 & 46.50 & 75.05 & 76.55 & 45.60 & \textbf{57.38} \\
        \midrule
        \multirow{2}*{\centering Mistral-v0.1 7B} 
        & 0\% & 42.60 & 46.57 & 29.80 & 73.80 & 79.55 & 53.92 & 83.70 & 82.10 & 44.00 & \textbf{59.56} \\
        & 20\% & 44.14 & 45.50 & 28.26 & 66.69 & 73.11 & 46.33 & 77.00 & 77.37 & 41.20 & \textbf{55.51} \\
        \bottomrule
    \end{tabular}}
    \caption{\textbf{Compressed models can retain 97\% of the zero-shot performance of the base model.} Zero-shot performances of the base non-compressed models (0\%) and the 20\% compressed models.}
    \label{tab:main_result}
\end{table*}

\begin{table*}
    \resizebox{\linewidth}{!}{%
    \begin{tabular}{cccc|ccccccc}
    \toprule
       \textbf{Model} & \textbf{Reduction}  & \textbf{Model Size} & \textbf{VRAM} &  s = 512 & s = 1024 &  s = 2048 & s = 4096 & s = 8192 & s = 16384 \\
    \midrule
    \multirow{2}*{\centering Phi-3 14B} 
       & 0\%      & 14B &  28 GB & 7171 t/s & 7216 t/s & 7197 t/s & 7057 t/s & 7010 t/s & 7036 t/s \\
       & 20\%     & 11B &  22.8 GB & 8622 t/s & 8686 t/s & 8509 t/s & 8349 t/s & 8268 t/s & 8269 t/s \\
    \midrule
       \multirow{2}*{\centering Mixtral-8x7B-v0.1 47B} 
       & 0\%      & 47B & OOM & OOM & OOM & OOM & OOM & OOM & OOM \\
       & 20\%     & 37B &  73.8 GB & 6515 t/s & 8143 t/s & 8444 t/s & OOM & OOM & OOM \\
    \midrule
       \multirow{2}*{\centering Phi-2 3B} 
       & 0\%      & 2.8B  &  6.8 GB & 29949 t/s & 30895 t/s & 30184 t/s & 30403 t/s & 26636 t/s & 21499 t/s \\
       & 20\%     & 2.2B  &  5.7 GB & 32451 t/s & 34561 t/s & 34276 t/s & 32479 t/s & 30351 t/s & 23735 t/s \\
    \midrule
       \multirow{2}*{\centering Mistral-v0.1 7B} 
       & 0\%      & 7.2B  &  15.2 GB & 13385 t/s & 13537 t/s & 13650 t/s & 13265 t/s & 12603 t/s & 12399 t/s \\
       & 20\%     & 5.8B  &  12.6 GB & 15416 t/s & 15832 t/s & 15947 t/s & 15446 t/s & 14589 t/s & 14304 t/s \\
    \bottomrule
    \end{tabular}}
    \caption{\textbf{Compressed models save memory and speedup computation.} Inference speed was measured on a single A100 GPU, using the whole test split of Wikitext2 using a batch size of 4 and by varying the sequence length from 512 to 16384 tokens. All the models were loaded in bf16 and used Flash-Attention2.}
    \label{tab:mem-speed}
\end{table*}

\begin{table*}[t]
    \resizebox{\textwidth}{!}{%
    \setlength{\tabcolsep}{3pt}
    \begin{tabular}{lcccccccccccc}
        \toprule
        \textbf{Model} & \textbf{TruQA} & \textbf{SIQA} & \textbf{LogiQA} & \textbf{WinoG} & \textbf{ARC-E} & \textbf{ARC-C} & \textbf{BoolQ} & \textbf{PIQA} & \textbf{OBQA} & \textbf{Average} \\
        \midrule 
        StableLM-2 1.6B & 38.90 & 48.52 & 26.73 & 63.30 & 68.39 & 38.57 & 74.80 & 76.93 & 39.00 & \textbf{52.79} \\
        Qwen-1 1.8B & 38.09 &  45.19 & 31.80 & 59.12 & 58.33 & 34.98 & 65.93 & 73.23 & 33.60 & \textbf{48.92} \\
        Qwen-2 1.5B &  45.93 & 45.85 & 31.18 & 66.22 & 60.56 & 36.09 & 72.26 & 75.35 & 36.40 & \textbf{52.20} \\
        \textbf{lillama} (Phi-2 1.7B) & 44.08 & 47.80 & 25.65 & 68.27 & 63.22 & 38.82 & 76.02 & 72.36 & 39.20 & \textbf{52.82} \\
        \bottomrule
    \end{tabular}}
    \caption{\textbf{40\% compressed Phi-2 3B competes with models of similar size while being compressed using only 13M tokens.} Compressing Phi-2 3B to 1.7B (by removing 40\% of the parameters) and comparing its zero-shot performance to other recent language models of similar sizes.}
    \label{tab:smollm}
\end{table*}

Computing the optimal compression rank of each matrix individually may be
costly: while search algorithms can be a solution for smaller models, they can be difficult to scale
to LLMs with billions of parameters. 
Assuming a target compression rate $N$, we propose to consider
three simple strategies: \textbf{uniform}, \textbf{top}, and \textbf{bottom} first compression strategies, which are illustrated in Appendix~\ref{sec:3strat}.

\myparagraph{Uniform.} This approach removes
$N$\% parameters to each weight matrix in the model. This strategy
involves training all LLM weights.

\myparagraph{Bottom.} This approach removes
$N$\% parameters from the model by prioritizing lower layers as
detailed in Algorithm~\ref{alg:two}. The core of the algorithm is line 7,
where the layers are ordered in bottom-top order and ranks are sorted in
decreasing order. The argument $k$ controls the \textit{spread} of compression: the
lower its value the more weights in the bottom layer will be compressed first.
The algorithm is also illustrated in Steps 1 and 2 in
Figure~\ref{fig:overview}. This strategy is efficient as only a subset of
weights in the bottom layers will be trained. Consequently, since there is no
need to forward through the whole model, only a subset of the weights needs to be loaded into GPU memory, making this approach scalable to larger models.

\myparagraph{Top.} This approach removes $N$\%
parameters from the model by prioritizing top layers. This consists roughly of
reversing the order of layers in the line 7 of Algorithm~\ref{alg:two}.
However, this strategy requires loading all the model's parameters into GPU
memory. It also involves forwarding through the entire model, even if only the
top layer weights are being trained.

\begin{algorithm}[h]
    {\small{
\caption{Bottom Layers First Compression Algorithm}\label{alg:two}
\KwIn{$\mathcal{M}$ is the base model and $\mathcal{M}'$ its copy; $S$ is the target size of the compressed model; $k$ is the minimum possible rank to set; $m$ the increment when generating ranks.}
\KwResult{The low-rank model $\mathcal{M}'$.}
$\mathcal{R} \gets \varnothing$ \tcp*[r]{Stack that will contain ranks and weight matrices.}

\For{each weight matrix $W \in \mathbb{R}^{d_{1} \times d_{2}}$ in $\mathcal{M}$}{
    $b \gets \min(d_{1}, d_{2})$\\
    \For{$r = k;\ r \leq b;\ r = r + m$}{
        \If{$r \times (d_{1} + d_{2}) < d_{1} \times d_{2}$}{
            $\mathcal{R} \gets \mathcal{R} \cup \{(r, W)\}$
        }
    }
}
\textbf{sort} $\mathcal{R}$ primarily by layer index in increasing order and secondarily by rank in decreasing order.\\
\While{$|\mathcal{M'}| > S$}{
    \textbf{get} the next $(r, W)$ from the stack $\mathcal{R}$ \\
    \textbf{compute} $A, B$ from $W$ using Eq. \ref{eq:svd} with the rank $r$\\
    \textbf{replace} $W$ by $A, B$
}
\Return{$\mathcal{M}'$}
}}
\end{algorithm}

\section{Experiments on Transformers}
\label{sec:xp}
\subsection{Setup}
\myparagraph{Models.} 
We first evaluate our method on various transformer-based LLMs: a 47B mixture-of experts language model (Mixtral-v0.1~8x7B~\citep{jiang_mixtral_2024}), medium-sized LLMs (Mistral-v0.1~7B~\citep{jiang2023mistral} and Phi-3~14B~\footnote{{\scriptsize{\url{https://huggingface.co/microsoft/Phi-3-medium-4k-instruct}}}}),
and a small language model (Phi-2~3B\footnote{{\scriptsize{\url{https://huggingface.co/microsoft/phi-2}}}}).

\myparagraph{Data.} For calibration data, we use 13 million tokens randomly
sampled from
Slim-Orca~\citep{SlimOrca}\footnote{{\scriptsize{\url{https://huggingface.co/datasets/Open-Orca/SlimOrca}}}},
an open-source replication of the Orca instruction
dataset~\citep{mukherjee2023orca}. In preliminary experiments, we also tested
other datasets, such as RedPajama \citep{together2023redpajama}, but found the results were better with instruction data.

\myparagraph{Hyperparameters.}
When compressing models using Algorithm~\ref{alg:two}, we set a minimum rank of $k=1024$ (except for Phi-3 14B, where $k=1536$ yielded better results). For all models, the rank increment $m$ is set to 256. We use the Teacher+Student loss (Figure~\hyperref[fig:teacher_student]{2c}). Section \ref{ssec:analysis} ablates these choices.

\myparagraph{Evaluation.} 
We evaluate the speed by measuring the time to forward a batch of 4 prompts of varying sequence lengths. Using lm-evaluation-harness~\citep{eval-harness}, we evaluate the zero-shot performance of the base and compressed models on 9 downstream tasks: TruthfulQA (\textbf{TruQA};
\citep{lin-etal-2022-truthfulqa}), Social IQa (\textbf{SIQA};
\citep{sap2019social}), \textbf{LogiQA} \citep{liu2020logiqa}, WinoGrande
(\textbf{WinoG}; \citep{sakaguchi2019winogrande}), Arc Easy (\textbf{ARC-E};
\citep{arc}), Arc Challenge (\textbf{ARC-C}; \citep{arc}), \textbf{BoolQ}
\citep{boolq}, \textbf{PIQA} \citep{piqa}, OpenBookQA (\textbf{OBQA};
\citep{obqa}).

\subsection{Results}

\myparagraph{Our method can retain 97\% of the zero-shot performance.} As shown in Table~\ref{tab:main_result}, models compressed by 20\% maintain over 93\% of the base models' performance, regardless of the size. The larger models, Phi-3 14B and Mixtral-8x7B-v0.1 47B, retain 97\% and 96\% of the performance, respectively. Overall, the models tend to lose the most performance on the ARC-C task, with losses also in the commonsense reasoning task WinoG. We provide additional results for 25\% and 30\% compression ratios in Appendix \ref{ssec:severe}.

\myparagraph{Low-Rank models are lighter and faster.}
Table~\ref{tab:mem-speed} shows the memory gain (VRAM) and speedup of the
compressed models. We can see that 20\% compressed models are lighter and up to
20\% faster. Notably, while Mixtral-8x7B-v0.1 cannot fit into a single A100-80GB
GPU, after compression, it can fit and process up to a 2048
context length with a batch size of 4. This is possible thanks to the memory
efficiency of the bottom-first compression strategy and to the local gradient
updates approach.

\myparagraph{Our approach can efficiently create Small Language Models.} We compressed Phi-2 3B to 1.7B (40\% of parameter reduction) and compare its performance to recent state-of-the-art small language models of equivalent size: StableLM-2~1.6B and Qwen-2~1.5B. Results are presented in Table~\ref{tab:smollm}.

\myparagraph{Compressed Models Show Good Recovery with Fine-Tuning.} Table \ref{tab:ft_result} demonstrates that fine-tuning these compressed models leads to performance recovery. Notably, the 40\% compressed Mistral 7B-v0.1 retains 91\% of the original performance despite being fine-tuned on only 191 million tokens for fine-tuning. Additional data could likely lead to even greater improvements.

\begin{table*}[t]
    \resizebox{\textwidth}{!}{%
    \setlength{\tabcolsep}{3pt}
    \begin{tabular}{cccccccccccc}
        \toprule
        \textbf{Model} & \textbf{Reduction} & \textbf{TruQA} & \textbf{SIQA} & \textbf{LogiQA} & \textbf{WinoG} & \textbf{ARC-E} & \textbf{ARC-C} & \textbf{BoolQ} & \textbf{PIQA} & \textbf{OBQA} & \textbf{Average} \\
        \midrule 
        \multirow{3}*{\centering Phi-2 3B} 
        & 0\% & 44.40 & 55.42 & 30.57 & 76.16 & 78.16 & 54.18 & 83.21 & 79.11 & 51.20 & \textbf{61.38} \\
        & 40\% & 44.08 & 47.80 & 25.65 & 68.27 & 63.22 & 38.82 & 76.02 & 72.36 & 39.20 & \textbf{52.82} \\
        & 40\% {\scriptsize FT} & 45.43 & 49.49 & 31.80 & 70.09 & 60.86 & 37.29 & 67.68 & 73.23 & 42.00 & \textbf{53.10} \\
        \midrule
        \multirow{3}*{\centering Mistral-v0.1 7B} 
        & 0\% & 42.60 & 46.57 & 29.80 & 73.80 & 79.55 & 53.92 & 83.70 & 82.10 & 44.00 & \textbf{59.56} \\
        & 40\% & 41.69 & 43.65 & 30.57 & 62.35 & 61.66 & 34.56 & 75.23 & 71.49 & 34.80 & \textbf{50.63} \\
        & 40\% {\scriptsize FT} & 41.60 & 52.25 & 29.65 & 66.61 & 66.75 & 39.93 & 79.39 & 74.37 & 38.29 & \textbf{54.32}\\
        \bottomrule
    \end{tabular}}
    \caption{\textbf{Compressed models recover well when finetuned.} Zero-shot performances of the base non-compressed models (0\%) and the 40\% compressed models, without and with fine-tuning (FT). Models were fine-tuned on the whole Slim-Orca dataset (191 million tokens).}
    \label{tab:ft_result}
\end{table*}

\begin{table*}[h!]
    \centering
    \resizebox{0.96\textwidth}{!}{%
    \setlength{\tabcolsep}{3pt}
    \begin{tabular}{lccccccccccc}
        \toprule
        \textbf{Model} & \textbf{Reduction} & \textbf{TruQA} & \textbf{SIQA} & \textbf{LogiQA} & \textbf{WinoG} & \textbf{ARC-E} & \textbf{ARC-C} & \textbf{BoolQ} & \textbf{PIQA} & \textbf{OBQA} & \textbf{Average} \\
        \midrule
        \multirow{2}*{\centering Falcon-Mamba 7B} 
        & 0\% & 53.40 & 52.66 & 31.34 & 74.66 & 81.73 & 58.96 & 83.12 & 81.56 & 48.60 & \textbf{62.89} \\
        & 20\% & 52.34 & 50.10 & 30.72 & 66.69 & 76.73 & 49.23 & 77.22 & 78.02 & 44.80 & \textbf{58.43} \\
        \midrule
        \multirow{2}*{\centering Mamba 3B} 
        & 0\% & 35.88 & 43.24 & 26.88 & 63.38 & 64.02 & 36.26 & 65.63 & 75.90 & 39.40 & \textbf{50.07} \\
        & 20\% & 37.50 & 42.43 & 27.04 & 61.40 & 62.16 & 35.58 & 64.89 & 73.29 & 40.00 & \textbf{49.37} \\
        \bottomrule
    \end{tabular}
    }
    \caption{\textbf{Mamba-based Language Models can be efficiently compressed}. Zero-shot evaluation results for the non-compressed (0\%) and for the compressed Mamba-based Language Models (20\%)}
    \label{tab:ssm}
\end{table*}

\label{ssec:analysis}
\begin{figure*}[h!]
    \centering
        \includegraphics[width=1.0\textwidth]{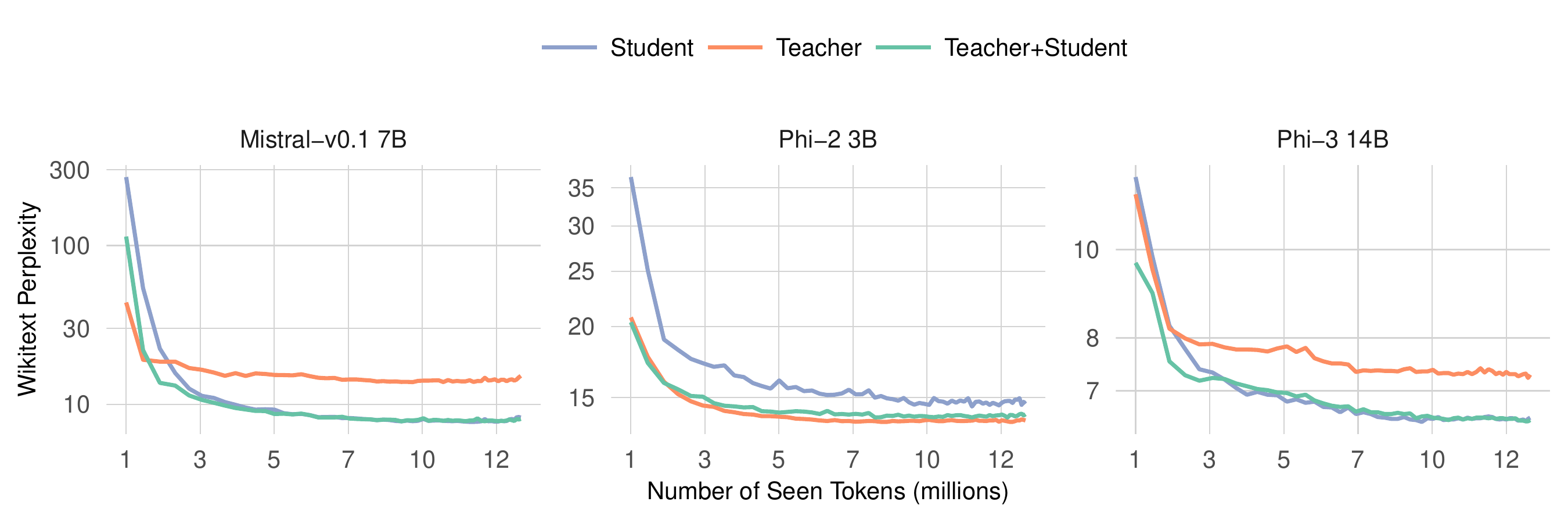}
        \caption{\textbf{The joint loss converges generally better}. Convergence of the three losses illustrated in Figure~\ref{fig:distillaton}, evaluated on the Wikitext2 test corpus perplexity during distillation.}
        \label{fig:losses}
\end{figure*}

\section{Experiments on Mamba Architecture}

In this section, we show the generalizability of our approach by evaluating it on the Mamba architecture for text
\citep{gu2024mambalineartimesequencemodeling}.
The attention mechanism in the Transformer is limited by its increasing complexity as a function of the input sequence length.
Therefore, Linear Attention Models have been proposed as alternatives to Transformers, in particular State Space Models, such as the Mamba architecture~\citep{gu2024mambalineartimesequencemodeling}, which have shown promising results.

\myparagraph{Method.} We tested our approach on two Mamba architectures: Mamba
3B~\citep{gu2024mambalineartimesequencemodeling} and Falcon-Mamba~7B\footnote{\scriptsize{\url{https://huggingface.co/tiiuae/falcon-mamba-7b}}}.
We compressed these models by 20\% using the same dataset and hyperparameters as in previous experiments with Transformers, with a minimum rank set to 1024 for both models.

\myparagraph{Results.} Table~\ref{tab:ssm} shows that, as for Transformers, Mamba-based Language Models can be efficiently compressed. Interestingly, the 20\% compressed Mamba 3B maintains 99\% of the performance. In Section \ref{ssec:fastmamba}, we also show the compressed Mamba models are lighter and faster.

\section{Analysis}

\begin{figure*}[ht]
    \begin{minipage}[c]{0.38\linewidth}
        \centering
        \includegraphics[width=\textwidth]{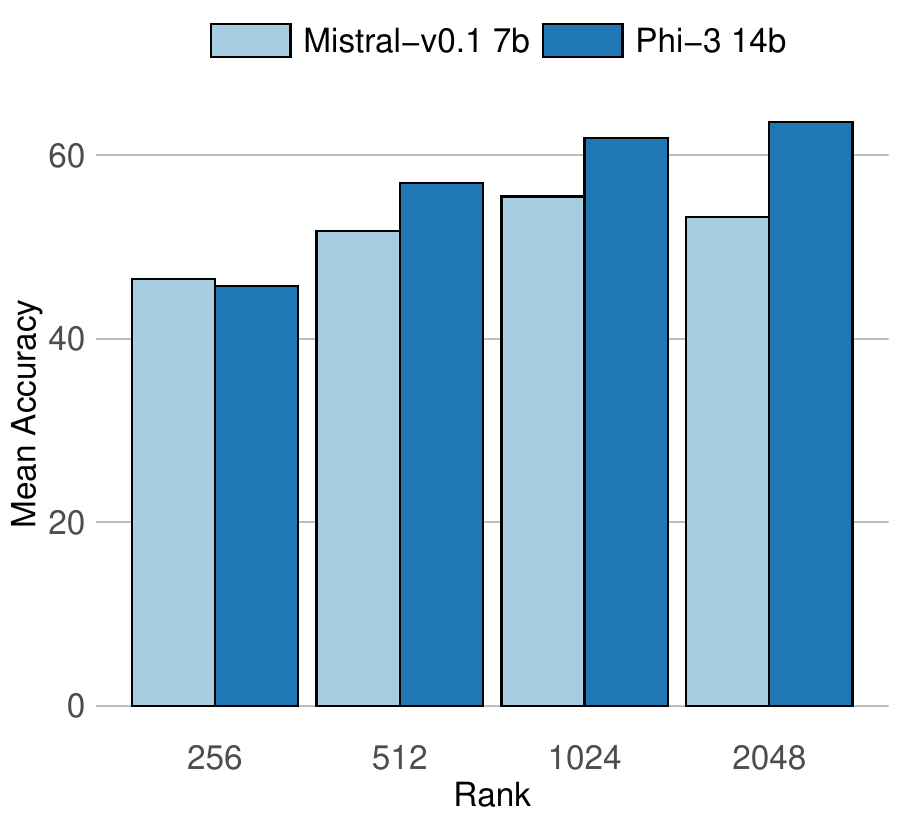}
        \caption{Mean accuracy over our test benchmarks as a function of the minimum rank $k$ in Algorithm~\ref{alg:two}}
        \label{fig:ranks}
    \end{minipage}
    \hfill
    \begin{minipage}[c]{0.58\linewidth}
        \begin{center}
            \includegraphics[width=\textwidth]{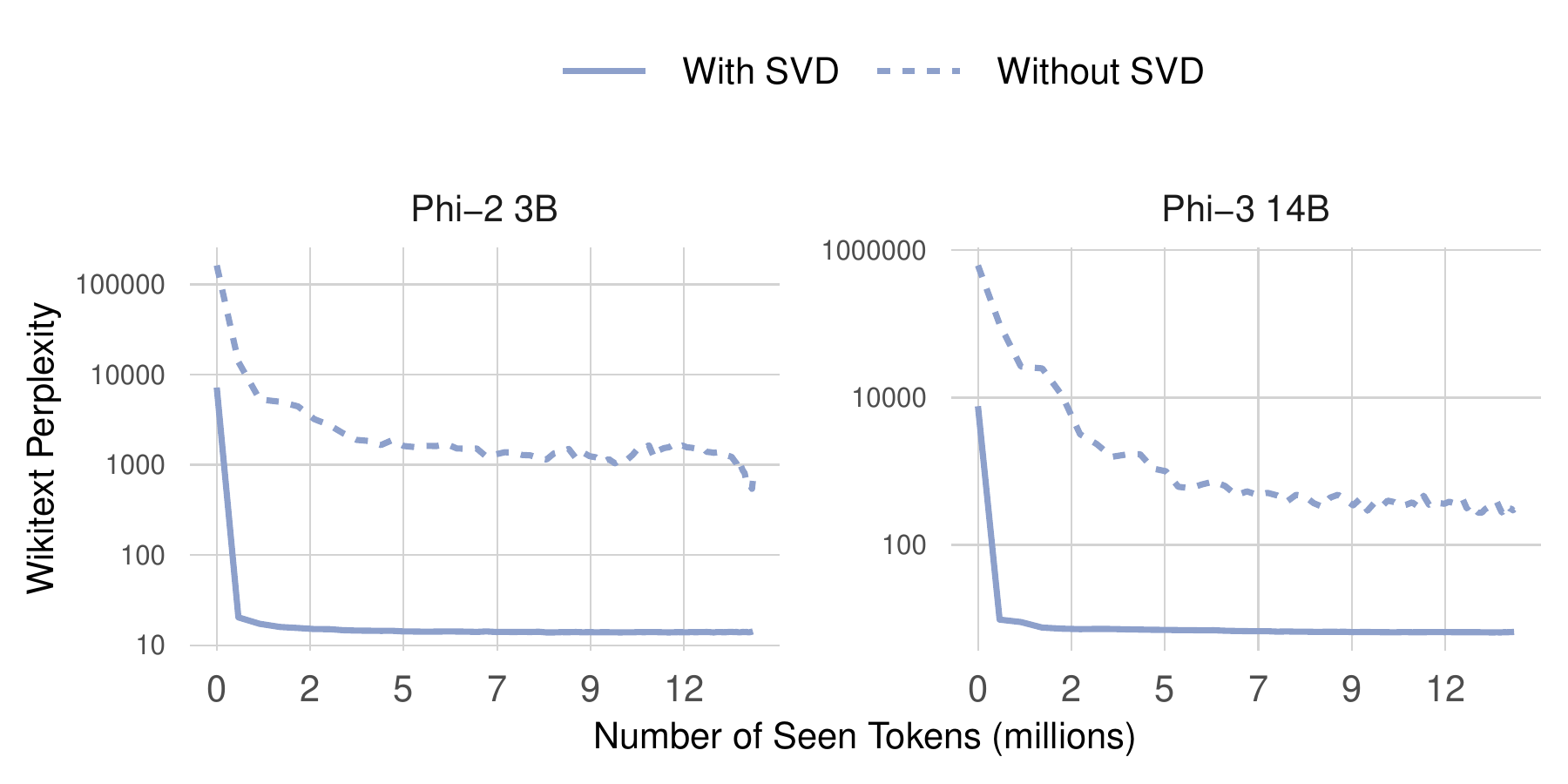}
            \captionsetup{width=\textwidth}
            \caption{Convergence when initializing low-rank weight randomly (without SVD) or with SVD.}
            \label{fig:w_wo_svd}
        \end{center}
    \end{minipage}
\end{figure*}

\begin{figure*}[ht]
    \begin{minipage}[c]{.56\textwidth}
        \fontsize{8.1pt}{8.1pt}\selectfont
        \centering
        \addtolength{\tabcolsep}{-0.55em}
        \begin{tabular}{cccccccc}
            \toprule
            \textbf{Reduction} & \textbf{Method} & \textbf{PIQA} & \textbf{WinoG} & \textbf{HellaSwag} & \textbf{ARC-E} & \textbf{ARC-C} & \textbf{Avg.} \\
            \midrule
            \multirow{2}{*}{0\%} 
            & SliceGPT & 79.11 & 75.77 & 73.83 & 78.32 & 54.18 & 72.24 \\
            & Reproduced & 79.11 & 76.01 & 73.60 & 78.41 & 54.35 & 72.30 \\
            \midrule
            \multirow{2}{*}{24\%} 
            & SliceGPT & 74.05 & 62.12 & 53.31 & 67.26 & 39.42 & 59.23 \\
            & \textbf{lillama} & \textbf{74.76} & \textbf{69.77} & \textbf{60.60} & \textbf{72.10} & \textbf{47.10} & \textbf{64.86} \\
            \bottomrule
        \end{tabular}
        \captionof{table}{\textbf{Comparison of our method with SliceGPT on zero-shot task evaluation for Phi-2 3B.} We give the evaluation scores before compression (0\%), both as reported by SliceGPT and as reproduced by us. Both methods use the Alpaca calibration dataset for compression.}
        \label{tab:comparisons}
    \end{minipage}
    \hfill
    \begin{minipage}[c]{.38\textwidth}
        \fontsize{8.8pt}{8.8pt}\selectfont
        \centering
        \addtolength{\tabcolsep}{-0.50em}
            \begin{tabular}{cccc}
            \toprule
            \textbf{Model} & \textbf{Teacher} & \textbf{Student} & \textbf{Tea+Stu} \\ 
            \midrule
            Phi-2 3B & 57.38 & 56.44 & 57.38 \\
            Phi-3 14B & 62.54 & 63.20 & 63.36 \\
            Mistral-v0.1 7B & 51.25 & 55.46 & 55.51 \\ 
            Mixtral-8x7B 47B & 58.87 & 60.49 & 60.19 \\ 
            \midrule
            \textbf{Average} & 57.51 & 58.90 & \textbf{59.11} \\
            \bottomrule
            \end{tabular}
        \captionof{table}{\textbf{The joint loss generally performs better}: Zero-shot performances of the thee losses using the same setup for each model, as described in Section~\ref{sec:xp}.}
        \label{tab:losses}
    \end{minipage}
\end{figure*}

\myparagraph{Which distillation loss to use?} Table~\ref{tab:losses} shows that the joint loss depicted in Figure~\hyperref[fig:teacher_student]{2c} generally produces the best results. Although the differences between the Student and Teacher+Student
losses may not seem significant, Figure~\ref{fig:losses} shows that the Teacher+Student loss leads to faster convergence than the Student loss alone. It also generally results in a lower final perplexity than using the Teacher loss only. We provide the raw values of the plot in Appendix \ref{ssec:logs_loss} to better analyze the differences.

\myparagraph{How to choose the rank values?}
In our bottom-first compression strategy, a small rank $k$ in
Algorithm~\ref{alg:two} compresses more severely the bottom layers and leaves
more upper layers uncompressed, hence saving GPU memory. However, a
small $k$ may also destroy knowledge and the model will struggle to recover
during the distillation phase. Figure \ref{fig:ranks} compares 4 rank
values for Mistral-v0.1~7B and Phi-3~14B and shows that higher ranks
lead to better performance, but are also more costly as more layers are distilled.

\myparagraph{Is SVD initialization necessary?} Figure~\ref{fig:w_wo_svd} shows
the perplexity curves for Phi-2 3B and Phi-3 14b when the low-rank matrices are
initialized with SVD or randomly. SVD initialization enables faster
convergence, with 8 million tokens being sufficient, i.e., roughly
4k examples of 2048 tokens.

\myparagraph{Cost.} All models in Section~\ref{sec:xp} are compressed
in less than one hour on a single A100 GPU. The precise cost
depends on the number of layers distilled determined by
the compression strategy (bottom, top, or uniform) and the minimum rank $k$ of
Algorithm~\ref{alg:two}: see Appendix~\ref{ssec:strategies} for details.

\myparagraph{Comparisons.} Table \ref{tab:comparisons} shows that our approach outperforms SliceGPT \citep{ashkboos_slicegpt_2024}, one-shot compression method for LLMs.

\section{Conclusion}
We propose a new efficient compression method that 
requires far less calibration data than most state-of-the-art approaches
and provides competitive performance for various models (Dense and MoE Transformers, Mamba), modalities (Text, Speech\footnote{See Appendix \ref{ssec:speech}}).
We show that combining SVD initialization with a joint Teacher and Student loss for local distillation enables fast convergence, and a bottom-up layer selection approach enables cost-effective compression.
In future work, we plan to study the complementary of our method with
quantization, and explore how the performance of
compressed models improves with continued pretraining.
\section{Limitations}
{\bf{Compatibility with quantization:}}
Although we propose in this work a pruning/compression method, the LLM size reduction
approach that is the most used nowadays is quantization. Both paradigms are theoretically
complementary, as removing parameters and quantizing the remaining ones is possible.
However, each step removes part of the information stored in the model, and finding the optimal
balance between the types of information that are removed with one or the other method is 
an important requirement that needs to be addressed before the proposed method may be adopted
at scale in common LLM libraries.
However, solving this challenge is difficult and beyond the scope of this work, and it certainly deserves
a dedicated study. \\
\\
{\bf{Compromise between specificity and genericity:}}
When designing an efficient LLM compression method like the one proposed in this work
and when trying to minimize the computational cost of this method, we inevitably have to make choices with
regard to important compromises, in particular how much can we improve the speed of our method
vs. how generic and applicable to a variety of conditions is our approach.
We have focused in this work on proposing a method that privileges genericity concerning raw speed, and this is why we do neither exploit GPU-specific kernels nor 2:4 and 4:8 semi-structured sparsity patterns. The question of efficiency also relates to the carbon cost of our proposed approach, which  depends on many factors: for instance,
the fact that our method works on CPU as well as on GPU makes it possible to reuse existing heritage hardware and not rely on carbon-costly GPU; also, in the lifecycle of an LLM, the additional cost incurred by compressing the model may be compensated by the future reduced cost when deployed.


\newpage

\onecolumn
\tableofcontents
\twocolumn
\newpage

\appendix
\section{Appendix}
\subsection{On bottom, top, and uniform compression strategies}
\label{ssec:strategies}
All the experiments presented here use the same hyperparameters as in Section \ref{sec:xp}.
\\
\\
\myparagraph{Differences in running time and memory.} The bottom-first compression strategy allows to compress all models, including Mixtral-8x7B-v0.1 47B. While the top-first strategy can compress Phi-3 14B, the uniform compression strategy fails to compress Phi-3 14B and Mixtral-8x7B-v0.1 (47B). Table~\ref{tab:ctime} gives the distillation time for each strategy and each model.
\\
\\
\myparagraph{Differences in performances.} Table~\ref{tab:cacc} presents the mean accuracies for each compression strategy. The bottom-first compression strategy performs well and is also more memory efficient. The top-first strategy does not improve accuracy while having higher time and memory complexity compared to the bottom-first approach.

\begin{table}[h]
    {\small{
    \begin{tabular}{lccc}
    \toprule
    \textbf{Model }& \textbf{Bottom} & \textbf{Top} & \textbf{Uniform} \\ 
    \midrule
    Phi-2 3B & 25min  & 32min  & 50min  \\ 
    Mistral-v0.1 7B & 26min  & 46min & 105min  \\ 
    Phi-3 14B & 30min & 50min  & OOM  \\ 
    Mixtral-8x7B-v0.1 47B & 47min & OOM  & OOM  \\ 
    \bottomrule
    \end{tabular}
    }}
    \caption{Compression times for the three compression strategies at 20\% parameter reduction on a single A100 GPU. For the bottom and top first compression strategies, we used the same hyperparameters as in Section \ref{sec:xp}.}
\label{tab:ctime}
\end{table}

\begin{table}[h]
    {\small{
    \begin{tabular}{lccc}
    \toprule
    \textbf{Model }& \textbf{Bottom} & \textbf{Top} & \textbf{Uniform} \\ 
    \midrule
    Phi-2 3B & 57.38  & 55.71  & 58.21  \\ 
    Mistral-v0.1 7B & 55.51  & 55.63 & 54.25  \\ 
    Phi-3 14B & 63.59 & 60.80  & OOM  \\ 
    Mixtral-8x7B-v0.1 47B & 60.19 & OOM  & OOM  \\ 
    \bottomrule
    \end{tabular}
    }}
    \caption{Average zero-shot performance for the three compression strategies at 20\% compression. For the bottom and top first compression strategies, we used the same hyperparameters as in Section \ref{sec:xp}.}
\label{tab:cacc}
\end{table}

\subsection{Compression at 20\%, 25\%, and 30\%}
\label{ssec:severe}
To show how performance is affected by different compression ratios, we compressed the models Phi-2 3B, Phi-3 14B, and Mistral-v0.1 7B at 20\%, 25\%, and 30\%. For the 20\% compression ratio, we used the same hyperparameters as described in Section \ref{sec:xp}. For compression ratios >20\%, we compressed Phi-3 14B using the bottom-first compression strategy with a minimum rank of $k=2048$. For Phi-2 3B and Mistral-v0.1 7B, we applied the uniform compression strategy. As shown in Table~\ref{tab:perf_morratios}, our method remains robust under severe compression ratios, with both Phi-3 14B and Phi-2 3B compressed at 30\% retaining 93\% of their base performance.

\begin{table*}[t]
    \resizebox{\textwidth}{!}{%
    \setlength{\tabcolsep}{3pt}
    \begin{tabular}{cccccccccccc}
        \toprule
        \textbf{Model} & \textbf{Reduction} & \textbf{TruQA} & \textbf{SIQA} & \textbf{LogiQA} & \textbf{WinoG} & \textbf{ARC-E} & \textbf{ARC-C} & \textbf{BoolQ} & \textbf{PIQA} & \textbf{OBQA} & \textbf{Average} \\
        \midrule 
        \multirow{4}*{\centering Phi-3 14B} 
        & 0\% & 57.63 & 57.06 & 37.94 & 76.01 & 81.36 & 61.60 & 88.56 & 81.34 & 50.40 & \textbf{65.77} \\
        & 20\% & 57.88 & 50.26 & 34.10 & 72.53 & 83.54 & 59.22 & 85.32 & 80.30 & 49.20 & \textbf{63.59} \\
        & 25\% & 56.37 & 51.38 & 32.26 & 70.56 & 80.98 & 57.51 & 83.88 & 79.76 & 47.00 & \textbf{62.19} \\
        & 30\% & 55.32 & 50.41 & 31.18 & 70.32 & 78.70 & 55.55 & 84.10 & 76.99 & 45.00 & \textbf{60.98} \\
        \midrule
        \multirow{4}*{\centering Phi-2 3B} 
        & 0\% & 44.40 & 55.42 & 30.57 & 76.16 & 78.16 & 54.18 & 83.21 & 79.11 & 51.20 & \textbf{61.38} \\
        & 20\% & 46.04 & 52.56 & 29.80 & 71.51 & 72.81 & 46.50 & 75.05 & 76.55 & 45.60 & \textbf{57.38} \\
        & 25\% & 44.55 & 51.64 & 30.88 & 71.35 & 71.93 & 45.99 & 79.79 & 74.43 & 43.20 & \textbf{57.08} \\
        & 30\% & 44.12 & 51.59 & 30.41 & 70.72 & 69.99 & 44.37 & 78.17 & 74.54 & 43.20 & \textbf{56.35} \\
        \midrule
        \multirow{4}*{\centering Mistral-v0.1 7B} 
        & 0\% & 42.60 & 46.57 & 29.80 & 73.80 & 79.55 & 53.92 & 83.70 & 82.10 & 44.00 & \textbf{59.56} \\
        & 20\% & 44.14 & 45.50 & 28.26 & 66.69 & 73.11 & 46.33 & 77.00 & 77.37 & 41.20 & \textbf{55.51} \\
        & 25\% & 40.52 & 44.93 & 30.11 & 68.11 & 65.07 & 40.61 & 78.44 & 73.99 & 40.20 & \textbf{53.55} \\
        & 30\% & 40.08 & 44.37 & 30.57 & 66.93 & 65.53 & 40.61 & 77.98 & 73.18 & 39.00 & \textbf{53.14} \\
        \bottomrule
    \end{tabular}}
    \caption{Zero-shot evaluation results for the base non-compressed models (0\%), and the 20\%, 25\%, and 30\% compressed models.}
    \label{tab:perf_morratios}
\end{table*}

\subsection{Convergence when using Teacher, Student or Teacher+Student loss}
\label{ssec:logs_loss}

As shown in Figure~\ref{fig:losses}, the Teacher+Student loss combines the best of both worlds: fast convergence due to the Teacher activations and high performance at inference thanks to the Student activations. We illustrate this in Table~\ref{tab:logs} for the first 2M training tokens. It shows that the Teacher+Student loss converges faster than the Student-only loss. While the Teacher-only loss also converges fast, it reaches a higher plateau and does not converge beyond that point, as shown in Figure~\ref{fig:losses}.

\begin{table}[h]
    {\scriptsize{
    \begin{tabular}{lccccc}
    \toprule
    \textbf{Model} & \textit{Seen Tokens} & \textbf{Teacher} & \textbf{Student} & \textbf{Tea+Stu} \\ 
    \midrule
    \multirow{5}{*}{Phi-2 3B} 
                              & 0       & 7185.31  & 7185.31  & 7185.31  \\
                              & 532480  & 20.74    & 36.56    & 20.34    \\
                              & 1056768 & 17.69    & 25.11    & 17.26    \\
                              & 1559879 & 16.00    & 18.97    & 15.92    \\
                              & 2000506 & 15.21    & 18.19    & 15.51    \\
    \midrule
    \multirow{5}{*}{Mistral-v0.1 7B} 
                              & 0       & 12365.44 & 12365.44 & 12365.44 \\
                              & 532480  & 43.79    & 270.16   & 114.02   \\
                              & 1056768 & 19.07    & 53.90    & 21.90    \\
                              & 1579121 & 18.56    & 22.40    & 13.64    \\
                              & 2056609 & 18.55    & 15.72    & 13.11    \\
    \midrule
    \multirow{5}{*}{Phi-3 14B} 
                              & 0       & 46362.82 & 46362.82 & 46362.82 \\
                              & 532480  & 9.77     & 12.94    & 8.89     \\
                              & 1056768 & 8.33     & 8.76     & 7.99     \\
                              & 1580799 & 7.70     & 7.85     & 7.44     \\
                              & 2069488 & 7.61     & 7.77      & 7.34     \\
    \bottomrule
    \end{tabular}
    }}
    \caption{Comparing the Wikitext perplexity convergence of the three losses for 2M training tokens.}
    \label{tab:logs}
\end{table}

\subsection{Speed up for Mamba architectures}
\label{ssec:fastmamba}

Table~\ref{tab:mem-speed-mamba} shows that compressed Mamba models are also lighter and faster than their original model.
\begin{table*}
    \centering
    \resizebox{\linewidth}{!}{%
    \setlength{\tabcolsep}{3pt}
    \begin{tabular}{ccccc|ccccc}
    \toprule
       \textbf{Model} & \textbf{Reduction} & \textbf{Model Size} & \textbf{VRAM} & s = 512 & s = 1024 & s = 2048 & s = 4096 & s = 8192 & s = 16384 \\
    \midrule
    \multirow{2}*{\centering Falcon-Mamba 7B} 
       & 0\%      & 7.27B & 15.3 GB & 8725 t/s & 9346 t/s & 9390 t/s & 9695 t/s & 9908 t/s & 9486 t/s \\
       & 20\%     & 5.82B  & 12.6 GB & 10227 t/s & 10427 t/s & 10477 t/s & 10814 t/s & 11108 t/s & 10647 t/s \\
    \midrule
    \multirow{2}*{\centering Mamba 3B} 
       & 0\%      & 2.77B  & 6.7 GB & 17994 t/s & 18556 t/s & 18819 t/s & 19618 t/s & 19231 t/s & 18602 t/s \\
       & 20\%     & 2.22B  & 5.6 GB & 19013 t/s & 19993 t/s & 20378 t/s & 21218 t/s & 20808 t/s & 20100 t/s \\
    \bottomrule
    \end{tabular}}
    \caption{\textbf{Compressing mamba models saves memory and speeds up computation.} Inference speed was measured on a single A100 GPU, using the whole test split of Wikitext2 with a batch size of 4 and by varying the sequence length from 512 to 16384 tokens. All models were loaded in bf16 and used Flash-Attention2.}
    \label{tab:mem-speed-mamba}
\end{table*}

\subsection{Activations are more low-rank than weights}
\label{ssec:srank}
\begin{figure*}[h!]
    \centering
        \includegraphics[width=1\textwidth]{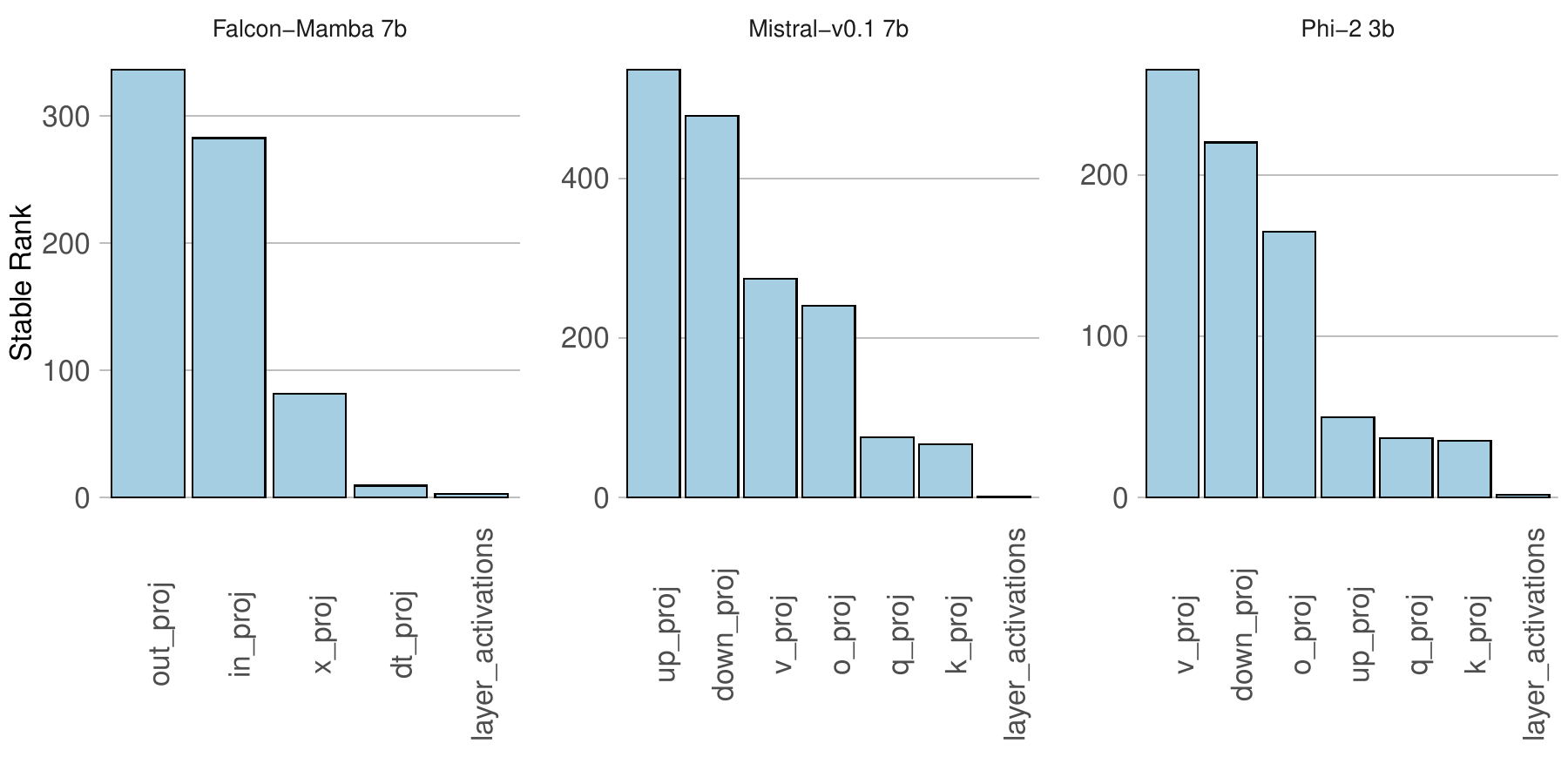}
    \caption{\textbf{Activations are low-rank.} Comparison of the stable rank of weights (\texttt{*\_proj}) and layer activations (\texttt{layer\_activation}). Each bar is the average stable rank across all layers.
    \label{fig:srank}}
\end{figure*}

We use stable rank ($srank$) as a proxy measure for rank: 
$${\text srank}(X) = \frac{\sum_{i=1}^{r} \sigma_i^2}{ \sigma_1^2}$$
with $X \in \mathbb{R}^{d_{1} \times d_{2}}$ being the input matrix (activations or weights), $r = \min(d_{1}, d_{2})$, and $\sigma_1 \geq \dots \geq \sigma_r$ the singular values of $X$.
Figure~\ref{fig:srank} shows the stable rank of various weight matrices and layer activations for three models: Falcon-Mamba, Mistral-v0.1 7b and Phi-2 3b. For all models, activations are significantly lower rank than weight matrices, which confirm previously published results~\cite{yu_compressing_2023}. This also explains why naive SVD solely doesn't work.

\subsection{Ease of Implementation of our approach}

\begin{figure*}[h!]
    \centering
        \includegraphics[width=0.8\textwidth]{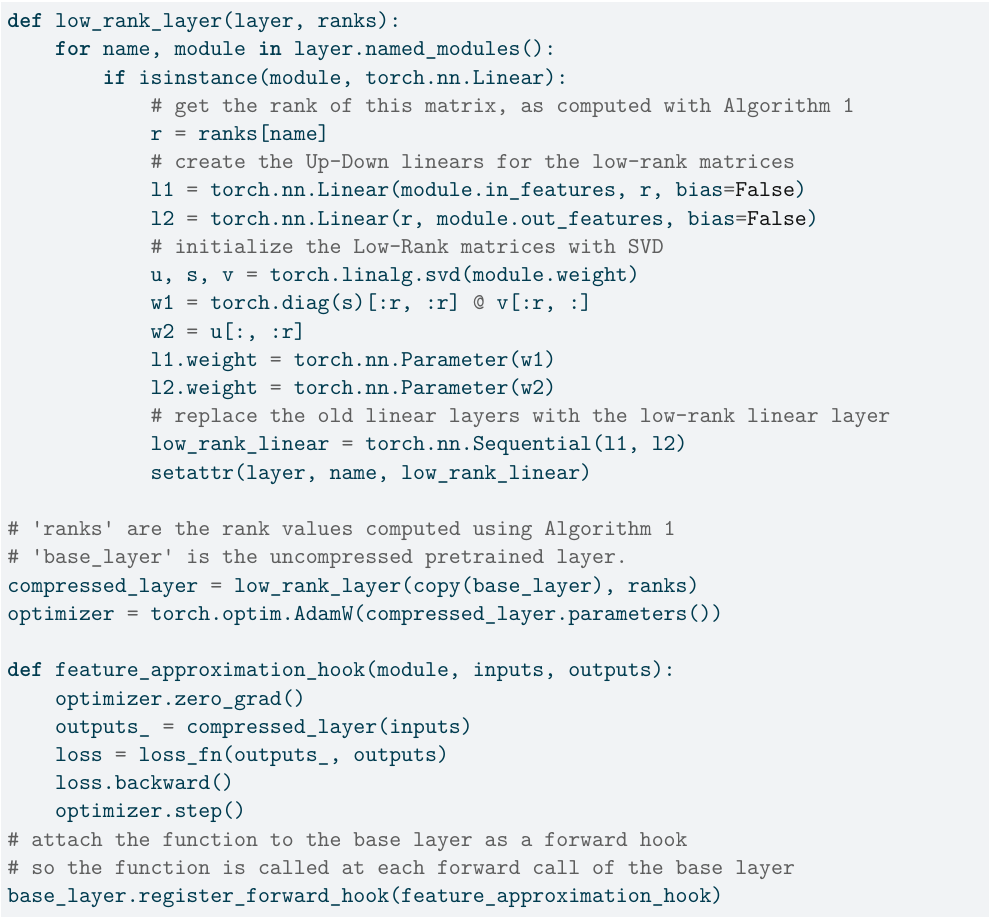}
        \caption{An example of PyTorch implementation of our approach with the Teacher loss (see Figure~\hyperref[fig:teacher]{2a}).}
        \label{fig:pseudocode}
\end{figure*}

Our proposed pruning approach gives good performance, is cost-efficient, supports various model architectures and modalities, and is further straightforward to implement.
Figure~\ref{fig:pseudocode} shows a simple pseudo-code implementation that can be easily added into most recent PyTorch models.

\subsection{Example of generated texts}
We also evaluated the compressed models by observing the text they generated. In general, at 20\% compression of Phi-2 3B, the generation ability and knowledge remained nearly intact. At 40\% compression, the model still retained its general knowledge and generation abilities but tended to be more repetitive and verbose at times. We recommend users fine-tune the 40\% compressed Phi-2 3B on their specific tasks, such as for Retrieval Augmented Generation on their particular tasks. Table~\ref{tab:sj} and \ref{tab:cw} give some generated texts from 20\% and 40\% compressed Phi-2 3B.

\begin{table*}
\footnotesize
\centering
\begin{tabular}{ p{4.5cm}  | p{4.5cm}  | p{4.5cm} }
\toprule
Base Model & -20\% & -40\% \\
\midrule
    Steve Jobs was a young man who had a passion for technology and design. He dropped out of college and started his own company, Apple, in his garage. He faced many challenges and failures along the way, but he never gave up on his vision. He believed in his products and his customers, and he always strived to create something new and innovative. He was not afraid to take risks and to challenge the status quo. He was a leader who inspired others to follow his example. He was a visionary who changed the world. &  Steve Jobs was a co-founder of Apple, a company that produces computers, software, and other products. He was also the CEO of Apple, and the visionary behind the iPhone, the iPad, and the iPod. He was known for his innovation, creativity, and leadership. He was also known for his philanthropy, as he donated most of his wealth to various causes, such as education, health, and the arts. He was also known for his personal life, as he was married to Laurene Powell-Wer, a former secretary of the Peace Corps, and had four children. He died in 2011 at the age of 56, after suffering from pancreatic cancer.

Questions:

- What was Steve Jobs' occupation?
- What were some of the products that Apple produced?
- What were some of the causes that Steve Jobs donated to?
- How many children did Steve Jobs have?
- When and how did Steve Jobs die?

Answers:

- Steve Jobs' occupation was a co-founder of Apple and the CEO of Apple.
- Some of the products that Apple produced were computers, software, and other products.
- Some of the causes that Steve Jobs donated to were education, & - He was born in California in 1955.

- He dropped out of college and started working at a computer company.

- He became the co-founder of Apple.

- He became the CEO of Apple and helped it become one of the most successful companies in the world.

- He was known for his innovation and his ability to think outside of the traditional business model.

- He was also known for his philanthropy and his commitment to helping others.

- He died in 2011.
\\
\bottomrule
\end{tabular}
\caption{Steve Jobs Story by the 20\% and 40\% compressed Phi-2 3B models, without \textit{any} finetuning.
{\bf{Prompt: "Here the story of Steve Jobs:"}}}
\label{tab:sj}
\end{table*}

\begin{table*}
\footnotesize
\centering
\begin{tabular}{ p{4.5cm}  | p{4.5cm}  | p{4.5cm} }
\toprule
Base Model & -20\% & -40\% \\
\midrule
The Civil War was caused by a number of factors, including economic and social differences between the North and the South, as well as disagreements over states' rights and slavery. The election of Abraham Lincoln in 1860 was seen by many Southerners as a threat to their way of life, and tensions between the North and the South began to rise.
The war began in 1861 when Confederate forces attacked Fort Sumter in South Carolina. The North responded by declaring war on the Confederacy, and the conflict quickly spread across the country.
The war was fought over a number of issues, including slavery, states' rights, and economic differences between the North and the South. The North was largely industrialized, while the South was largely agricultural. This led to tensions over tariffs and other economic issues.
The war was also fought over the issue of slavery. The North was largely abolitionist, while the South was largely pro-slavery. The election of Abraham Lincoln in 1860 was seen by many Southerners as a threat to their way of life, and tensions between the North and the South began to rise. & The American Civil War was a conflict that took place from 1861 to 1865. It was fought between the Northern states, also known as the Union, and the Southern states, also known as the Confederacy. The main cause of the war was the issue of slavery. The Southern states wanted to keep slavery, while the Northern states wanted to abolish it. This led to tensions and eventually led to the war.

What were the major battles of the Civil War? Here is the story.

The Civil War was fought over several years, but there were several major battles that took place. The most famous battles were the Battle of Gettysburg, the Battle of Antieta, and the Battle of Bull Run. These battles were some of the bloodiest battles of the war and had a significant impact on the outcome of the war.

What was the Emancipation Proclamation? Here is the story.

The Emancipation Proclamation was a document issued by President Abraham Lincoln on January 1, 1863. It declared that all slaves in Confederate territory were to be set free. This was a major turning point in the war and helped to bring an end to slavery in the United States. & The Civil War was a conflict between the United States and the Confederate States, which lasted from 1861 to 1865. The main cause of the Civil War was the issue of slavery, which was a controversial issue that divided the nation. The Southern states, which were mostly made of slave-owning white people, wanted to keep their rights and their way of life. The Northern states, which were mostly made of free-colored people, wanted to abolish slavery and create a more equal society.

The Civil War began when the Confederate States declared their secession from the United States, which means they wanted to break away from the country. The United States responded by sending troops to the South, which started the war. The war lasted for four years, and it was one of the most important events in American history. The war ended with the defeat of the Confederate States, and the United States became a unified nation.

The Civil War had many consequences, both positive and negative. Some positive consequences were that the United States became more united and more democratic, and that the issue of slavery was finally resolved. Some negative consequences were that many people died, and that the war caused a lot of damage and destruction.
\\
\bottomrule
\end{tabular}
\caption{The Civil War Story by the 20\% and 40\% compressed models, without \textit{any} finetuning. {\bf{Prompt: "What is the cause of the Civil War? Here is the story."}} }
\label{tab:cw}
\end{table*}

\subsection{Speech modality: Whisper}
\label{ssec:speech}
We evaluate next our approach on a state-of-the-art speech model: {\it whisper-medium.en}~\citep{radford2022robustspeechrecognitionlargescale}, a 764M parameters encoder-decoder speech recognition model trained on English.

\myparagraph{Method.} We take inspiration from the \textit{deep encoder,
shallow decoder} approach~\citep{kasai2021deepencodershallowdecoder}, where the
authors observe that the decoder in encoder-decoder models does not need to be
very deep to achieve good performance. 
Therefore, we keep the encoder
unchanged and remove 70\% of the decoder's parameters. We then approximate the
low-rank matrices using the same method and hyperparameters described in
Section \ref{sec:xp}. However, since Whisper-medium.en is already a model with
less than 1 billion parameters, we apply a uniform compression strategy to
achieve a 70\% reduction in the decoder parameters.

\myparagraph{Data.} For calibration, we used 60\% of randomly sampled examples from Librispeech-train-100~\citep{librispeech}. We also fine-tuned the compressed model on a subset of 10 hours of speech randomly sampled from Librispeech-train-100. We froze the encoder parameters during this fine-tuning step and trained only the decoder.

\begin{table}[H]
    {\small{
    \begin{tabular}{ccccc}
    \toprule
        \textbf{Reduction} &\textbf{Model} & \textbf{WER ($\downarrow$)}  & {\bf VRAM} & \textbf{Speed} ($\uparrow$) \\ 
    \midrule
        0\% & 764M & 25.40 & 3.5GB & 2395t/s\\
        37\%  & 485M & 30.84 & 2.4GB & 2717t/s \\
        0\% {\scriptsize FT} & 764M & 9.20 & 3.5GB & 2395t/s\\
        37\% {\scriptsize FT} & 485M & \textbf{12.14 } & 2.4GB & 2717t/s \\ 
    \bottomrule
    \end{tabular}
    }}
    \caption{\textbf{The 37\% compressed Whisper is faster, 1.46$\times$ lighter and outperforms the base Whisper after fine-tuning on just 10 hours of speech. The non-finetuned compressed model retains over 82\% of its original performance.} The WER is computed on the English test subset of the Fleurs dataset. The GPU Memory (VRAM) and speed (tokens/second) correspond to transcription of 400 tokens from a dataset containing 2,048 speech examples, using a batch size of 128. A single A100 GPU is used and the model is load in FP32, without Flash-Attention.}
    \label{tab:wer}
\end{table}

\myparagraph{Evaluation results.} We evaluate the models' Word Error Rate (WER), memory footprint and latency on an
out-of-distribution test dataset: the test split of the English Fleurs dataset~\citep{fleurs}.
Table~\ref{tab:wer} gives the WER for compressed and uncompressed models. We can see that a 37\% compressed whisper maintains 82\% of the performance. Further fine-tuning this compressed model on only 10 hours of speech decreases the WER, even outperforming the base model. This compressed model is 31\% lighter and 14\% faster.

\subsection{Illustration of the 3 compression strategies}
\label{sec:3strat}

\begin{figure}[h!]
    \centering
        \includegraphics[width=0.5\textwidth]{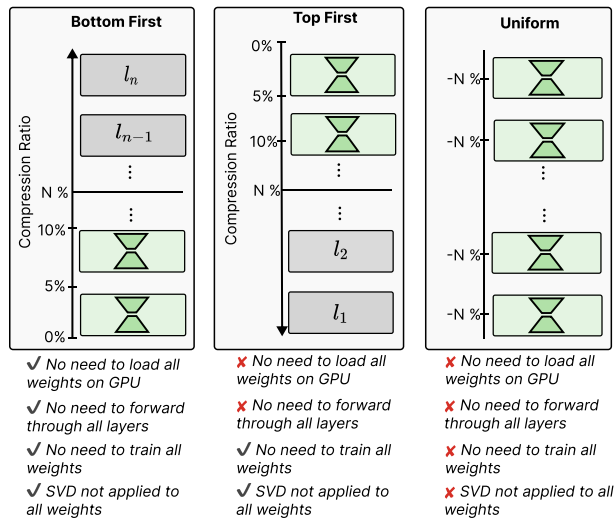}
    \caption{Illustration of 20\% compression when using the three strategies: bottom, top and uniform. Green boxes are compressed layers with their low-rank matrices while gray boxes are the non-compressed layers. $l_{1}$, $l_{2}$, ..., $l_{n-1}$, $l_{n}$ are the layer indexes.}
    \label{fig:compression}
\end{figure}

Figure~\ref{fig:compression} intuitively illustrates the three compression strategies evaluated in our method:
uniform, bottom-first and top-first, explaining their respective advantages and drawbacks.

\subsection{General hyper-parameters}
\label{sec:hypparms}

The following general hyper-parameters are used in all our experiments, except when stated otherwise.

\myparagraph{Implementation.}
We implement our approach using version 4.44.2 of the Huggingface Transformers library~\citep{wolf2020huggingfaces} and PyTorch version 2.4.0+cu121. For all experiments, we use the AdamW optimizer with a learning rate of $8.6e^{-4}$. Note that since our approach is local, each student layer has its own optimizer. This also allows us to perform local gradient updates and avoid storing the entire large PyTorch computation graph in memory. All experiments were conducted on a single A100 GPU, regardless of the model.

\myparagraph{Hyperparameters.} When compressing the models using Algorithm~\ref{alg:two}, we set a minimum rank of $k=1024$ for all models except Phi-3 14B, for which we found higher ranks work well, so we set $k=1536$. For all models, we set $m=256$ for the increment to generate ranks.
We apply the Teacher+Student loss, as illustrated in Figure~\hyperref[fig:teacher_student]{2c}, to all models, as we found that this loss converges faster and generally performs better. Additionally, we use the bottom-first compression strategy, which proved to be effective and less memory-intensive than other strategies. Section \ref{ssec:analysis} ablates these choices. To measure inference speed, we measure the time to forward a batch of prompts
through the entire model and present the results in tokens per second.

\subsection{Metrics used when evaluating with lm-evaluation-harness}
We provide in Table~\ref{tab:task_metric_map} the metrics we used to evaluate the models in Section \ref{sec:xp}.
\begin{table}[h]
\centering
    \begin{tabular}{ll}
    \toprule
    \textbf{Task} & \textbf{Metric} \\ 
    \midrule
    arc\_challenge   & acc\_norm \\ 
    arc\_easy        & acc\_norm \\ 
    piqa             & acc\_norm \\ 
    social\_iqa      & acc \\ 
    logiqa           & acc\_norm \\ 
    truthfulqa\_mc2  & acc \\ 
    winogrande       & acc \\ 
    boolq            & acc \\ 
    openbookqa       & acc\_norm \\ 
    \bottomrule
    \end{tabular}
\caption{Benchmarks and corresponding metrics that were used to evaluate the models in Section \ref{sec:xp}.}
\label{tab:task_metric_map}
\end{table}

\subsection{A Tiny Model for a Low-Resource Language}
\label{sec:hausa}

For most human languages, there is not enough data (generally less than 1B
tokens) to pretrain large LLMs. In such cases, data-efficient approaches are
preferred. We experimented compressing a tiny language model for Hausa, a
low-resource language for which there is not billions of tokens available for
continued pretraining. We used InkubaLM~\citep{tonja2024inkubalm}, a 422M
parameter language model designed for five low-resource languages. Since 60\%
of the parameters are in the embeddings, we focused on compressing the input
embedding $W_{e}$ and the prediction head $W_{o}$. This is easily achieved with
our approach by locally distilling the low-rank embeddings and prediction head:

\[
\widehat{\Delta W} = \underset{{\Delta W}}{\mathrm{argmin}} \;\; \mathcal{L}(Wx, \Delta Wx)
\]
where $\Delta W$ are the low-rank embedding matrices or prediction head, and $x$ is an input example.

We used a rank of 1024, compressing the model by 30\%. Then, we applied a
lightweight local distillation using our approach, with 64k randomly sampled
Hausa sentences from InkubaMono\footnote{\href{https://huggingface.co/datasets/lelapa/Inkuba-Mono}{https://huggingface.co/datasets/lelapa/Inkuba-Mono}} and the Aya-Dataset~\citep{singh2024aya}. We compared the models with MobiLLaMA~\citep{thawakar2024mobillama} and
SmoLLM\footnote{\href{https://huggingface.co/blog/smollm}{https://huggingface.co/blog/smollm}} using the Afrimmlu benchmark~\citep{afrimmlu}. As shown in Table~\ref{tab:inkuba}, the compressed InkubaLM model for Hausa retains 93\% of its base performance.

\begin{table}[h!]
    \centering
    {\small{
    \begin{tabular}{lcc}
    \toprule
    \textbf{Model } & \textbf{Model Size (Billion)} & \textbf{Accuracy}  \\ 
    \midrule
    SmoLLM & 1.7 & 21.80  \\ 
    MobiLLaMA & 1.3 & 21.40  \\  
    \midrule
    InkubaLM-base & 0.422 & 29.20 \\
    InkubaLM-30\% & 0.299 & 27.40  \\ 
    \bottomrule
    \end{tabular}
    }}
    \caption{Zero-shot performance on afrimmlu for the 30\% compressed InkubaLM model for the Hausa language.}
\label{tab:inkuba}
\end{table}

\end{document}